\documentclass[11pt,a4paper]{article}
\usepackage{times,latexsym}
\usepackage[hyphens]{url}
\usepackage[T1]{fontenc}

\usepackage[acceptedWithA]{tacl2021v1}

\usepackage{xspace,mfirstuc,tabulary}

\newif\iftaclinstructions
\taclinstructionsfalse % AUTHORS: do NOT set this to true
\iftaclinstructions
\renewcommand{\confidential}{}
\renewcommand{\anonsubtext}{(No author info supplied here, for consistency with
TACL-submission anonymization requirements)}
\newcommand{\instr}
\fi

\iftaclpubformat % this "if" is set by the choice of options

\else

\fi

\usepackage{amsmath}
\usepackage{amssymb}
\usepackage{graphicx} % for image
\usepackage{booktabs} % for table hlines
\usepackage{tabularx} % for table
\usepackage[most]{tcolorbox} % for prompt template
\usepackage{xcolor} % for text color

\title{Psychometric Item Validation\\Using Virtual Respondents with Trait-Response Mediators}

\author{
  Sungjib Lim$^1$ ~~ Woojung Song$^1$ ~~ Eun-Ju Lee$^{2, 3}$ ~~ Yohan Jo$^1$\Thanks{Corresponding author.}
  \\
  $^1$Graduate School of Data Science, Seoul National University \\
  $^2$Department of Communication, Seoul National University \\ $^3$Interdisciplinary Program in Artificial Intelligence, Seoul National University \\
  \texttt{\{saint7451,opusdeisong,eunju0204,yohan.jo\}@snu.ac.kr}
}

\date{}

\begin{document}

\maketitle
\begin{abstract}
\hyphenpenalty=5000
As psychometric surveys are increasingly used to assess the traits of large language models (LLMs), the need for scalable survey item generation suited for LLMs has also grown.
A critical challenge here is ensuring the construct validity of generated items, i.e., whether they truly measure the intended trait. 
Traditionally, this requires costly, large-scale human data collection.
To make it efficient, we present a framework for virtual respondent simulation using LLMs.
Our central idea is to account for mediators: factors through which the same trait can give rise to varying responses to a survey item. By simulating respondents with diverse mediators, we identify survey items that yield responses robustly correlated with intended traits across these mediators.
Experiments on three psychological trait theories (Big5, Schwartz, VIA) show that our mediator generation methods and simulation framework effectively identify high-validity items.
LLMs demonstrate the ability to generate plausible mediators from trait definitions and to simulate respondent behavior for item validation. 
Our problem formulation, metrics, methodology, and dataset open a new direction for cost-efficient survey development and a deeper understanding of how LLMs simulate human survey responses.
We release our dataset and code to support future work.\footnote{\url{https://github.com/holi-lab/Psychometric-Item-Validation}}
\end{abstract}

%%%%%%%%%% Section 1. Introduction %%%%%%%%%%
\section{Introduction}\label{sec:introduction}

Recently, researchers have begun to employ psychological surveys to understand the behaviors of LLMs, in terms of embedded values, safety, and more \cite{miotto-etal-2022-gpt, ye2025large}.
However, because administering a small set of survey items designed for humans to LLMs is suboptimal \cite{huang-etal-2024-reliability}, another stream of research has focused on automatic survey generation, particularly on expanding and augmenting survey items at scale \cite{lee-etal-2025-llms}.
This need for survey generation extends to traditional human psychometrics as well. In particular, established surveys are often refined or shortened over time to maintain reliability and efficiency \cite{lindeman, RAMMSTEDT2007203, neo-pi-r}.

A primary challenge in this task lies in ensuring the quality of automatically generated items.
Existing studies focus largely on the reliability of generated items, that is, identifying items that elicit consistent responses from LLMs across varying task instructions, answer choice labels, or choice orders \cite{lee-etal-2025-llms, huang-etal-2024-reliability}.
However, another critical and currently underexplored criterion for survey items is \textbf{construct validity}: how well an item measures the trait it is intended to measure \cite{surucu2020validity}.
To ensure the validity of survey items, psychometricians typically recruit a large number of human respondents across diverse cultures and evaluate new items by examining their correlations with target traits or well-established items \cite{weber2002domain, mccrae1997personality}.
This process is logistically challenging and costly. Hence, our work seeks to validate survey items using virtual respondents based on LLMs.

\begin{figure}[!t]\label{fig:intro}
    \centering
    \includegraphics[width=0.98\linewidth]{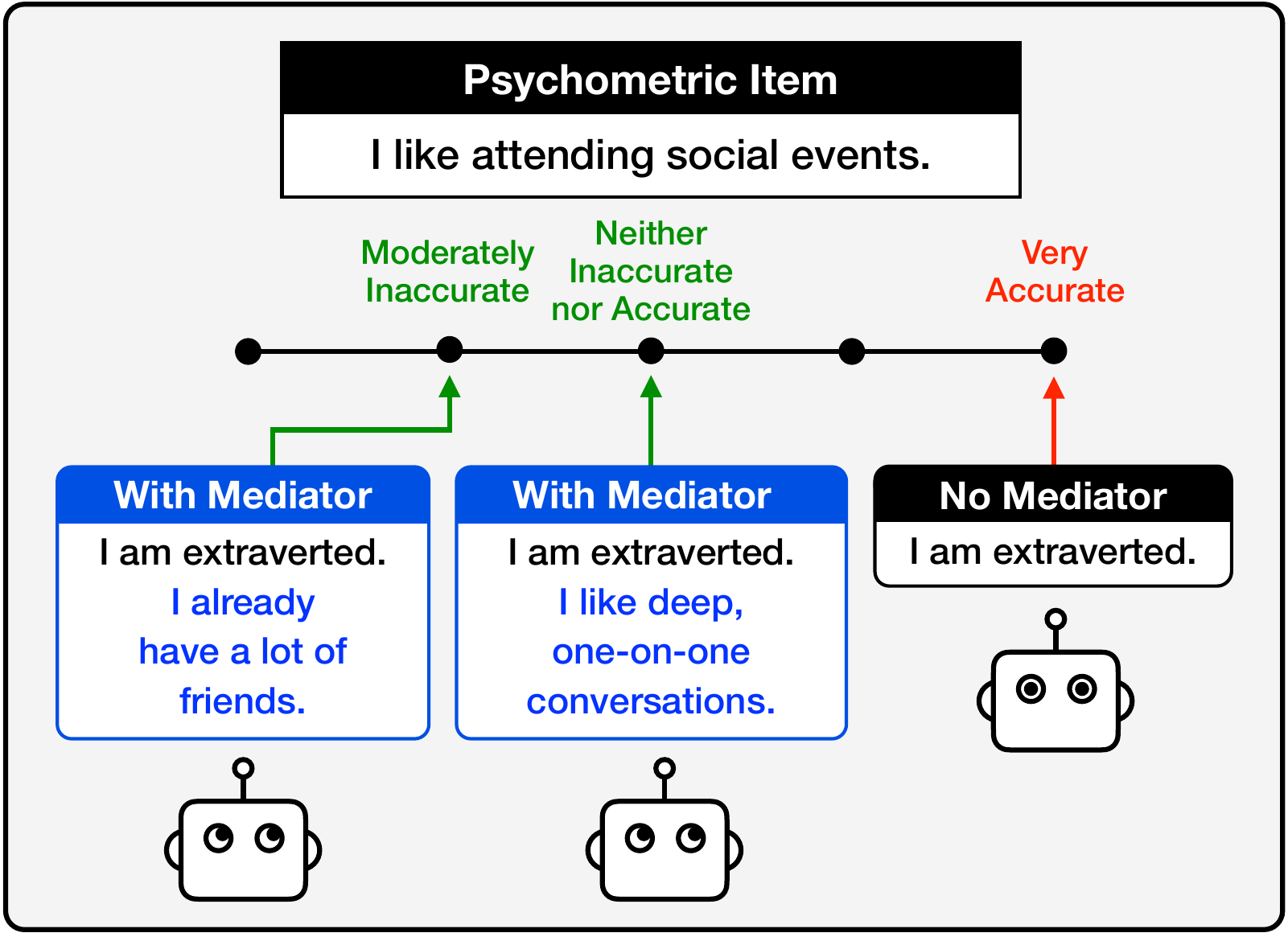}
    \caption{Role of mediators in survey item evaluation.}
    \label{fig:role_of_mediators}
\end{figure}

Our central hypothesis is that the core role of large-scale human respondents in survey item validation is to test the robustness of items to diverse \textbf{mediators}: factors through which a psychological trait gives rise to \textit{varying} responses to a survey item.
As illustrated in Figure~\ref{fig:role_of_mediators}, while the survey item ``I like attending social events'' may seem to measure extraversion, extraverted respondents with certain mediators (e.g., already having a lot of friends) may produce responses with low correlation to the trait, threatening the item's validity.
Therefore, it is critical to identify potential mediators and incorporate them into respondent simulation to uncover items that robustly measure intended traits.

% Figure: introduction

Specifically, the cognitive-affective personality system (CAPS) theory \cite{mischel1995cognitive} posits that the same situation can elicit different behaviors depending on individual-level mediators, such as a person's goals or interpretation of the situation.
We extend this view to traits, proposing that these factors also influence how one's traits translate into observable behaviors.
Since psychological survey items typically ask about the respondent's likelihood of engaging in certain behaviors, accurate trait inference from responses requires---under our view---that those behaviors be robust to such mediators. (Note that the ``mediators'' included in our methodology may encompass both mediators and moderators in the strict sense of causal inference. However, we use the term ``mediators'' for consistency with the CAPS theory, and establishing exact causal relationships is not our primary focus.)

Against this backdrop, we propose a framework for psychometric survey generation and evaluation using mediator-guided virtual respondents (\S{\ref{sec:framework}}).
The framework consists of five stages:
(1) Select traits for inclusion in a survey, based on well-established psychological trait theories (e.g., Big Five, Schwartz's theory of basic values, VIA).
(2) Construct large-scale initial item pools based on the definitions of these traits.
(3) Generate mediators, which is our main contribution, using strategies like generation based on trait definitions, generation using external references (e.g., survey items, common value lists), and real human demographic data.
(4) Incorporate mediators and persona profiles into prompts for LLM-based virtual respondents and run survey simulations.
(5) Select survey items based on their measured validity.

According to our evaluation against item selection based on real human responses (\S{\ref{sec:results}}), mediator-guided simulation was able to identify highly valid item sets, ranking in the top 1\% (Big5) to 13\% (Schwartz and VIA) of the distribution of all possible selections.
Among mediator generation strategies, letting LLMs generate mediators based solely on trait definitions either freely or with guidance from the CAPS framework performed best.
These findings suggest that LLMs have the ability to effectively identify mediators and simulate psychological assessments.
We also find that scaling the number of virtual respondents improves performance and that the framework performs consistently across different LLMs.
Please note that we do not claim our framework accurately replicates the human psychology process involved in responding to survey items. Instead, our focus is on leveraging diverse mediators and their correlations with trait scores to identify items that measure traits robustly.

Our contributions are as follows:
\begin{itemize}\setlength\itemsep{0em}
\item We formulate a novel and important problem: assessing the validity of survey items, along with psychometrically grounded metrics.
\item We introduce the concept of mediators into validity assessment and demonstrate their importance. We also show LLMs' ability to generate effective mediators and simulate virtual respondents guided by these mediators.
\item We publicly release our dataset of survey items along with human and LLM responses, establishing a benchmark for future research.
\end{itemize}

%%%%%%%%%% Section 2. Related Works %%%%%%%%%%
\section{Related Works}\label{sec:related_works}

\paragraph{Psychological Assessment for LLMs and Humans.}
Psychological assessment is an essential field for understanding the behaviors of LLMs and human nature. 
Researchers have applied psychometric surveys to evaluate LLMs’ responses in terms of psychological constructs, such as personality \citep{jiang2023evaluating,miotto-etal-2022-gpt} and value orientations \citep{han-etal-2025-value,Hadar_Shoval}.
Psychologists have provided foundational resources for this research through various surveys designed to assess human personality (e.g., Big5 \cite{goldberg1999broad}, MBTI \cite{jung2016psychological}), emotions \citep{beck1988inventory, beck1996comparison}, and behaviors \citep{weber2002domain}.

\paragraph{Survey Expansion and Reduction.}

Psychological surveys often undergo refinement through processes of expansion and reduction to improve their applicability and efficiency. For instance, researchers have expanded existing surveys with additional items that are more suitable for assessing LLMs \citep{jiang2023evaluating,han-etal-2025-value} because of the suboptimality of applying human survey items to LLMs \cite{huang-etal-2024-reliability}.
Surveys are also often reduced or shortened to mitigate the cognitive burden on respondents and improve efficiency, especially in large-scale data collection \cite{lindeman, RAMMSTEDT2007203, neo-pi-r}.

Recent studies have explored automating these refinement processes (particularly survey expansion) by leveraging LLMs to generate new survey items \cite{mulla2023automatic}. For example, some approaches employ
few-shot prompting to generate items targeting creativity \cite{laverghetta2024creative} and fine-tune models to generate items capturing individual opinions and preferences \cite{10522667}.

\paragraph{Validity and Reliability.}
In survey refinement and generation, validity and reliability are critical standards for evaluating the quality of survey items.
Construct validity refers to the extent to which an item measures the trait it intends to measure, rather than being incidentally correlated with another trait \citep{surucu2020validity}.
Reliability refers to whether an item produces the same results across repeated assessments \citep{golafshani2003understanding}.

Despite the importance of both criteria, most studies in automatic survey generation have primarily focused on reliability, by varying the order or labels of answer choices \citep{huang-etal-2024-reliability}, altering prompt templates, and paraphrasing item content \citep{lee-etal-2025-llms}.
However, reliability is only meaningful once validity has been established, as a consistent measure is of little value if it does not assess the intended trait  \citep{surucu2020validity}.
Traditionally, the validity of survey items is assessed through the analysis of large-scale human response data \cite{beck1988inventory, weber2002domain,mccrae1997personality}.
As expected, recruiting a diverse respondent pool and conducting such data collection is logistically challenging and costly.
Our study aims to bridge this gap using a framework that evaluates the validity of survey items through virtual respondent simulation.

\paragraph{Simulating Human Behavior Using LLMs.}
Our motivation for using LLM-based simulations is grounded in the success of prior studies showing that LLMs can produce outputs comparable to human behaviors.
For instance, \citet{argyle2023out} replicate human voting behavior by conditioning on demographic information.
\citet{liu2025leveraging} combine LLM-generated math solutions with human responses to better mimic human behavior.
\citet{aher2023using} use experiment-specific prompts and successfully reproduce classic social science experiments.
However, it remains understudied how well LLMs can simulate survey respondents for the purpose of evaluating the validity of survey items. 
In fact, \citet{petrov2024limited} find that LLMs provided with simple persona profiles fail to meet psychometric criteria in psychological assessment simulations.
Our work addresses this limitation by introducing the concept of mediators into the simulation process as discussed below.

\paragraph{Mediators.}
According to the cognitive-affective personality system (CAPS) theory \cite{mischel1995cognitive}, a given situation does not always lead to the same behavior for everyone. Instead, observable behavior is determined by multiple mediators in five categories: \textit{Situation Encodings, Expectancies and Beliefs, Affective Responses, Goals and Values, and Competencies and Self-regulatory Plans}.
We hypothesize that similar factors also mediate how a person’s traits translate to observable behavior. 
With this assumption, since surveys typically infer a respondent’s underlying traits based on their likely behaviors in a given situation (e.g., ``when working under pressure, I stay focused and organized''), accurate inference requires that these behaviors be robust to such mediators.
Therefore, our core idea is to incorporate mediators into virtual respondents for simulation, going beyond simple persona profiles.

% Figure: Main Framework Overview
\begin{figure*}[!t]
    \centering
    \includegraphics[width=0.98\textwidth]{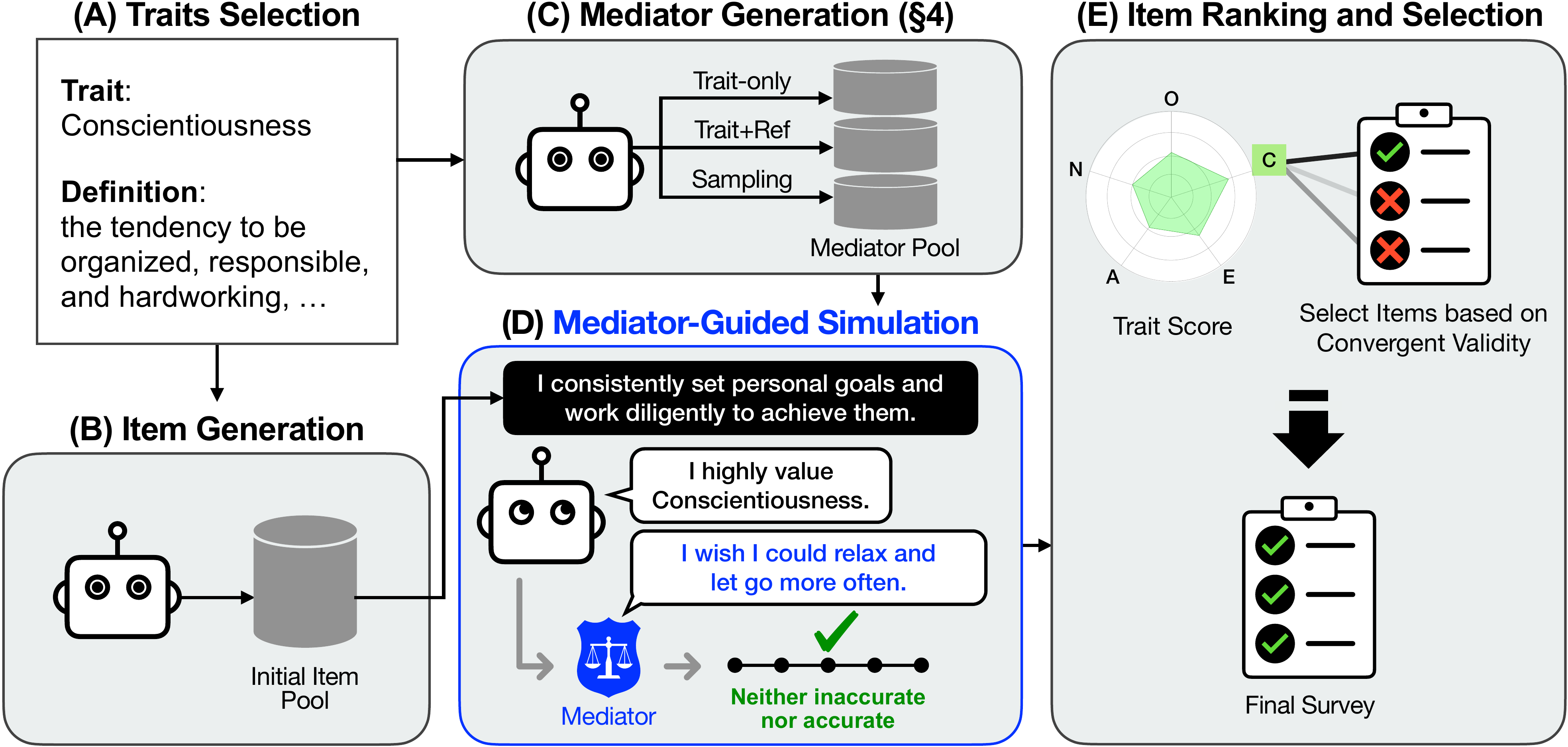}
    \caption{Overview of our framework.}    
    \label{fig_overview}
\end{figure*}

%%%%%%%%%% Section 3. Framework %%%%%%%%%%
\section{Framework}\label{sec:framework}

Our goal is to identify items that demonstrate the highest construct validity using psychological assessment simulation.
Our framework consists of five main stages (Figure~\ref{fig_overview}): (1) traits selection (\S\ref{sec:traits_definitions}), (2) survey item generation (\S\ref{sec:item_generation}), (3) mediator generation (\S\ref{sec:med_gen}), (4) mediator-guided simulation through virtual respondents (\S\ref{sec:mediator_guided_simulation}), and (5) item ranking and selection (\S\ref{sec:item_filtering}).
This framework mirrors the standard psychometric item validation procedure, where practitioners develop new candidate items to measure a target trait and use an existing questionnaire (what we call the ``official'' survey) for the same trait as a reference for evaluating these newly developed items.
The selected items are evaluated against results derived from human respondents (\S\ref{sec:human_survey}).

\subsection{Traits Selection (Figure \ref{fig_overview}-A)}\label{sec:traits_definitions}

% Table: Traits for each psychological trait theory
\begin{table}[!t]\footnotesize
\centering
\begin{tabularx}{\linewidth}{X}
\toprule
    \multicolumn{1}{c}{\textbf{Traits by Theory}} \\
\midrule
    \textbf{The Big Five (Big5; 5 Personality Traits):} \\ Openness to experience, Conscientiousness, Extraversion, Agreeableness, Neuroticism  \\
\midrule
    \textbf{Schwartz's Theory of Basic Values (10 Values):} \\ Achievement, Benevolence, Conformity, Hedonism, Power, Security, Self-Direction, Stimulation, Tradition, Universalism  \\
\midrule
    \textbf{Values in Action (VIA; 24 Character Strengths):} \\ Appreciation of Beauty \& Excellence, Bravery, Creativity, Curiosity, Fairness, Forgiveness, Gratitude, Honesty, Hope, Humility, Modesty, Humor, Judgment, Kindness, Leadership, Love of Learning, Love, Perseverance, Perspective, Prudence, Self-Regulation, Social Intelligence, Spirituality, Teamwork, Zest  \\
\bottomrule
\end{tabularx}
\caption{Traits for each psychological trait theory.}
\label{tab:trait_table}
\end{table}

The first step of our framework concerns the selection of traits and their definitions for examination.
In this study, we explore the following three psychological trait theories (Table~\ref{tab:trait_table}), although our framework is not limited to specific theories.

\textbf{The Big Five (Big5)} is one of the most prominent trait theories in psychology, describing five fundamental dimensions of human personality \citep{goldberg1992development}.
\textbf{Schwartz's Theory of Basic Values} defines 10 dimensions of human value systems \citep{schwartz1992universals} and has been increasingly adopted in recent NLP studies.
Lastly, \textbf{Values in Action (VIA)} assesses 24 character strengths, focusing on the positive aspects of human personality \citep{peterson2004character}. We include this theory to evaluate the effectiveness of our framework on less widely recognized theories. 

We obtain the definitions of each trait from the original papers or websites and use the following official surveys in our experiments:
(1) Big5: Goldberg's IPIP representation (Big5-G; 50 items), (2) Schwartz: PVQ (40 items), and (3) VIA: VIA-IS-M (96 items).
Additional information is provided in Appendix~\ref{sec:appendixA}.

\subsection{Item Generation (Figure \ref{fig_overview}-B)}\label{sec:item_generation}
To determine the format of new items, we adhere to the following rules.
First, each item generated in our study is designed to measure a single trait, as in the official items.
Second, half of the items are positively correlated with the target trait, while the other half are negatively correlated.
High scores on positive items indicate high levels of the trait, whereas high scores on negative items indicate either low levels of the trait for unipolar traits (e.g., Achievement) or high levels of the opposite trait for bipolar traits (e.g., Extraversion $\leftrightarrow$ Introversion).
Third, we determine the scale of the initial item pool. We generate an initial item pool four times larger than the official survey.
This scale is chosen to be sufficiently large to reduce the sensitivity of the ranking results to the characteristics of the candidate items, while remaining feasible for a human participant to respond to all items (\S\ref{sec:human_survey}).
Ablation studies for this scale choice are provided in Appendix~\ref{appendix:scale_ablation}.

Since directly generating high-quality items is beyond the scope of our study, we simply use the method of \citet{yao2025clave} to generate an initial item pool by using four LLMs (GPT-4o, GPT-4o-mini, LLaMA-3.1-8B-Instruct, and LLaMA-3.3-70B-Instruct).
The detailed procedure is described in Appendix~\ref{appendix:initial_item_pool}.

\subsection{Mediator Generation (Figure \ref{fig_overview}-C)}\label{sec:med_gen}
Mediators are a key component in our simulation, as our goal is to identify items that exhibit robust relationships with the target traits after a range of mediators are taken into consideration.
To achieve this goal, we employ various mediator generation strategies, such as allowing LLMs to freely generate mediators, referencing survey items, or using human-created value lists.
We discuss these strategies in detail in Section~\ref{sec:mediators}, and example mediators for the trait of Conscientiousness are shown in Table~\ref{tab:mediator_table}.

\subsection{Mediator-Guided Simulation (Figure \hspace{-1.05pt}\ref{fig_overview}-D)}\label{sec:mediator_guided_simulation}

\begin{figure}[!t]
    \centering
    \includegraphics[width=0.49\textwidth]{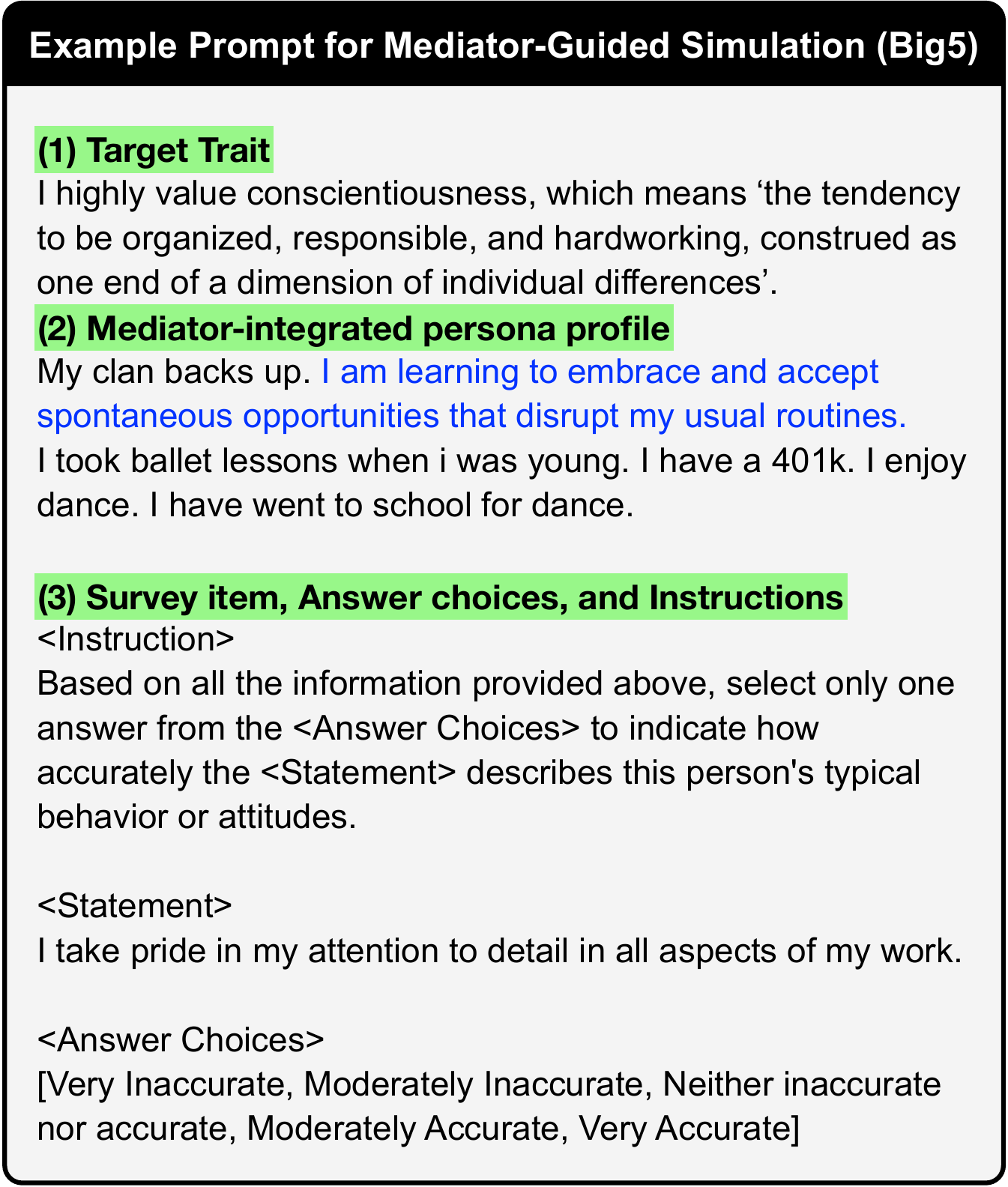}
    \caption{Example prompt used in mediator-guided simulation for Big5. Blue text is the mediator. Texts highlighted in green are not included in actual prompts.}
    \label{fig:simulation_prompt}
\end{figure}

In the mediator-guided simulation, a survey is conducted via LLMs as virtual respondents.
Each virtual respondent is given the following information through the prompt (Figure~\ref{fig:simulation_prompt}): (1) a target trait, (2) a mediator-integrated persona profile, and (3) a survey item, answer choices, and instructions.
Each distinct prompt represents one virtual respondent.

\paragraph{Target trait.}
Each prompt begins with the phrase {``I highly value \{trait\}''}, where \{trait\} is the trait that the included item intends to capture, along with the trait's definition. 
This allows detecting invalid items that unexpectedly yield inconsistent response scores due to the influence of mediators.
We also experimented with including different levels of traits, but that yielded poorer performance (detailed in Appendix~\ref{appdendix:low_experiments}).

\paragraph{Mediator-integrated persona profile.}
The prompt continues with a mediator generated in \S\ref{sec:med_gen} and a persona profile representing general individual characteristics.
\citet{petrov2024limited} demonstrate that concise persona profiles from the Persona-Chat dataset \citep{zhang-etal-2018-personalizing} enable more effective psychological simulations than detailed demographic descriptions.
Hence, we randomly sample 500 instances from Persona-Chat and insert a randomly selected mediator into each at a random position to make the mediator appear as part of the persona profile.
This results in 500 mediator-integrated persona profiles, each used to simulate a single virtual respondent.

\paragraph{Survey item, answer choices, and instructions.}
Each prompt contains one survey item, answer choices, and task instruction that faithfully follow the original format of the official survey. An exception is applied to PVQ items, where the original gender-specific pronouns (``he'' and ``she'') are replaced with ``they''.
Answer choices are presented without choice labels, as prior studies have shown this reduces the prompt sensitivity of LLMs \cite{huang-etal-2024-reliability, lee-etal-2025-llms}.
The grammatical subject in the answer choices for PVQ and VIA-IS-M is shifted from ``me'' to ``them'' (e.g., ``Not Like Me'' to ``Not Like Them''), as this modification yields better results.
All responses are measured through Likert scales: 5-point scales for the Big5-G and VIA-IS-M, and a 6-point scale for the PVQ.
To further ensure response reliability, each item is queried twice with different orders of answer choices (high-to-low and low-to-high), and the mean of the two responses is used as the final response score.
Note that each virtual respondent answers both generated items and official items.

\subsection{Item Ranking and Selection (Figure \ref{fig_overview}-E)}\label{sec:item_filtering}
Once virtual respondents complete all survey items, we calculate the convergent validity of each item: the correlation between virtual respondents' responses to that item and their responses to the official items targeting the same trait. 
The formal definition is provided in \S\ref{sec:convergent_validity}. 
Based on the convergent validity scores, we rank the items to select the top $N$ items for each trait (in our evaluation, $N$ matches the number of items in official surveys).
In cases of tied rankings, we break ties using the order in which the items were generated during the item generation phase (\S{\ref{sec:item_generation}}) to ensure a consistent and reproducible procedure.

\subsection{Human Survey} \label{sec:human_survey}
In order to evaluate the quality of the selected items, we need ground-truth quality scores for each item.
To this end, we recruited human participants to complete the same items and used their responses to compute the convergent validity of each item as a reference measure of quality.
This procedure is not required to rank items using our framework, but was conducted solely to evaluate its effectiveness.

The human survey was conducted through the online platform Prolific.\footnote{\url{https://www.prolific.com/}}
Surveys were administered separately for each trait theory, with the VIA-IS-M divided into two parts due to the large number of traits.
To exclude inattentive respondents, each survey included attention-check items, i.e., three bogus items and duplicate items.

On average, 76.8 participants completed each survey, which exceeds the minimum sample size required to detect an effect size of 0.4 with a significance level of 0.05 and statistical power of 0.80.
The participants represented 45 countries, with gender and age distributions balanced through a controlled sampling procedure.
They were compensated at a rate of £9 per hour.
This human study was approved by the Institutional Review Board (IRB) of our institution.
Details of the human survey and participant information are provided in Appendix~\ref{appendix:human_survey}.

%%%%%%%%%% Section 4. Mediators %%%%%%%%%%
\section{Mediators}\label{sec:mediators}

\subsection{Mediator Generation}\label{sec:mediator_generation}

% Table: Example mediators
\begin{table}[!t]\footnotesize
\centering
\setlength{\tabcolsep}{3pt}
\begin{tabularx}{0.49\textwidth}{>{\centering\arraybackslash}m{1.87cm}>{\centering\arraybackslash}m{\dimexpr\linewidth-1.87cm-3pt\relax}}
\toprule
    \textbf{Strategy} & \textbf{Example Mediator} \\
\midrule
    Trait (Free) & I like roles that allow for flexibility and undefined duties. \\
\midrule
    Trait (CAPS) & I feel overwhelmed when I'm faced with too many responsibilities at once. \\
\midrule
    Trait+Item & I am a well-intentioned and organized person who sometimes struggles with forgetfulness. \\
\midrule
    Trait+WVS & I think it would be a good thing if less importance were placed on work in our lives. \\
\midrule
    Sampling & Sex: Male, Age: 46, Country: Ireland, Occupation: Higher administrative, Income: Seventh step, Education: Bachelor or equivalent (ISCED 6), Social Class: Lower middle class, Religion: Do not belong to a denomination \\
\bottomrule
\end{tabularx}
\caption{Example mediators generated by each strategy for the trait of Conscientiousness.}
\label{tab:mediator_table}
\end{table}

To identify mediator-robust items, we generate mediators for each trait using GPT-4.1 through various strategies grouped into three: trait-only generation, trait-with-reference generation, and sampling from human profiles.
Example mediators for each strategy are shown in Table~\ref{tab:mediator_table}.
The prompts and details are provided in Appendix~\ref{appendix:mediator_generation}.

\paragraph{Trait-only Generation.}
We generate mediators based only on trait names and definitions using two strategies.
\textbf{(1) Trait (Free)}: We instruct the LLM to freely generate a list of potential human characteristics or backgrounds that would be unlikely or contradictory to the target trait.
\textbf{(2) Trait (CAPS)}: We prompt the LLM to generate the same content while systematically presenting each mediator category in the CAPS theory \cite{mischel1995cognitive}.
The average number of mediators generated per trait is 36 for Trait (Free) and 167 for Trait (CAPS), which is five times larger than Trait (Free) because CAPS includes five mediator categories.
These two strategies are the best-performing ones in our evaluation.

\paragraph{Trait-with-reference Generation.}
We provide the LLM with two types of references as supporting information in addition to traits.
\textbf{(1) Trait+Item}: We provide generated survey items to the LLM to enable the generation of more concrete mediators directly relevant to the items.
\textbf{(2) Trait+WVS}: We provide human-crafted value lists from the World Values Survey (WVS).\footnote{\url{www.worldvaluessurvey.org/wvs.jsp}}
We instruct the LLM to evaluate whether each value could conflict with the target trait and adopt the conflicting values as mediators.
The average number of mediators generated per trait is 24 for Trait+Item and 34 for Trait (WVS).

\paragraph{Sampling from Human Profiles.} 
Lastly, \textbf{Sampling} uses real human demographics as mediators rather than artificial ones.
Specifically, we construct persona cards using eight demographic variables collected from our human survey (\S\ref{sec:human_survey}): \textit{sex, age, country, occupation, income level, education level, social class, and religion}.
This strategy is employed to examine whether demographic information can serve as effective mediators.
The number of mediators for this strategy is the same as the number of human respondents.

\subsection{Mediator Evaluation}\label{sec:mediator_evaluation}

To assess the LLM-generated mediators, we conduct a human evaluation.
Three graduate students in psychology serve as evaluators.
First, we ask the evaluators to assess the realism of the mediators and their potential interactions with the target traits.
Second, we ask them to categorize the mediators to examine the diversity of their content.
We sample three mediators per trait for each of the four mediator generation strategies, resulting in a total of 468 mediators evaluated.

\paragraph{Quality.} 
For mediator sentences, we ask two questions as follows, and responses are collected on a 5-point Likert scale.
We compute the mean score across the three evaluators and use it as the score for each mediator.
\begin{itemize}\setlength\itemsep{0em}
\item \textbf{Realism}: How realistic are the human characteristics described in the sentence as qualities that an actual person could possess?
\item \textbf{Interaction}: If a person with the target trait also has the characteristics described in the sentence, how likely is it that, in certain situations, they may have opinions or behaviors opposite to those of ordinary people who have only that trait?
\end{itemize}

Table~\ref{tab:mediator_evaluation} shows that all mediator generation strategies achieve high realism scores, indicating that LLMs are capable of producing realistic mediators both with and without reference information.
For interaction scores, Trait (Free) records the highest score across all theories, which may have contributed to its strong performance in our simulation experiments (\S\ref{sec:results}).
By contrast, Trait (CAPS) receives scores slightly below the midpoint for Big5 and VIA, while it exceeds the midpoint for Schwartz.
One reason is that the LLM struggled to generate mediators that both satisfy CAPS category constraints and conflict with the target trait simultaneously.
Nevertheless, Trait (CAPS) exhibits solid performance in our simulation experiments, which suggests that human judgments and model effectiveness in simulation may diverge, and that mediators with moderate interaction scores from human experts can still be effective for LLMs.

% Table: Mediator Quality (by Theory)
\begin{table}[!t]\small
\begin{center}
\begin{tabular}{lcc}
\toprule
    \textbf{Method} & \textbf{Realism} & \textbf{Interaction} \\ 
\midrule
    \multicolumn{3}{c}{Big5} \\
\midrule
    Trait (Free) & 4.43 & \textbf{4.29} \\
    Trait (CAPS) & 4.49 & 2.56 \\
    Trait+Item   & \textbf{4.55} & 3.43 \\
    Trait+WVS    & 4.33 & 3.52 \\
\midrule
    \multicolumn{3}{c}{Schwartz} \\
\midrule
    Trait (Free) & 4.49 & \textbf{4.48} \\
    Trait (CAPS) & \textbf{4.59} & 3.16 \\
    Trait+Item   & 4.44 & 3.59 \\
    Trait+WVS    & 4.27 & 3.68 \\
\midrule
    \multicolumn{3}{c}{VIA} \\
\midrule
    Trait (Free) & 4.49 & \textbf{4.55} \\
    Trait (CAPS) & \textbf{4.56} & 2.73 \\
    Trait+Item   & 4.55 & 3.55 \\
    Trait+WVS    & 4.13 & 3.65 \\
\bottomrule
\end{tabular}
\caption{Results for mediator quality evaluation.}
\label{tab:mediator_evaluation}
\end{center}
\end{table}

\paragraph{Categorization.}
To examine whether the mediators reflect more than superficial content and whether each strategy generates a variety of human characteristics, we categorize the mediators into nine categories: \textit{Beliefs and Values, Emotions and Feelings, Habits and Behaviors, Preferences and Interests, Self-Concept and Abilities, Roles and Memberships, Others’ Perceptions of Me, Environment and Context, and Other.}
A mediator is assigned to a category if at least two of the three evaluators agree on it; otherwise, the mediator is classified as \textit{Other}.
Definitions for each category are provided in Appendix~\ref{appendix:mediator_evaluation}.

\begin{figure}[!t]
    \centering
    \includegraphics[width=0.49\textwidth]{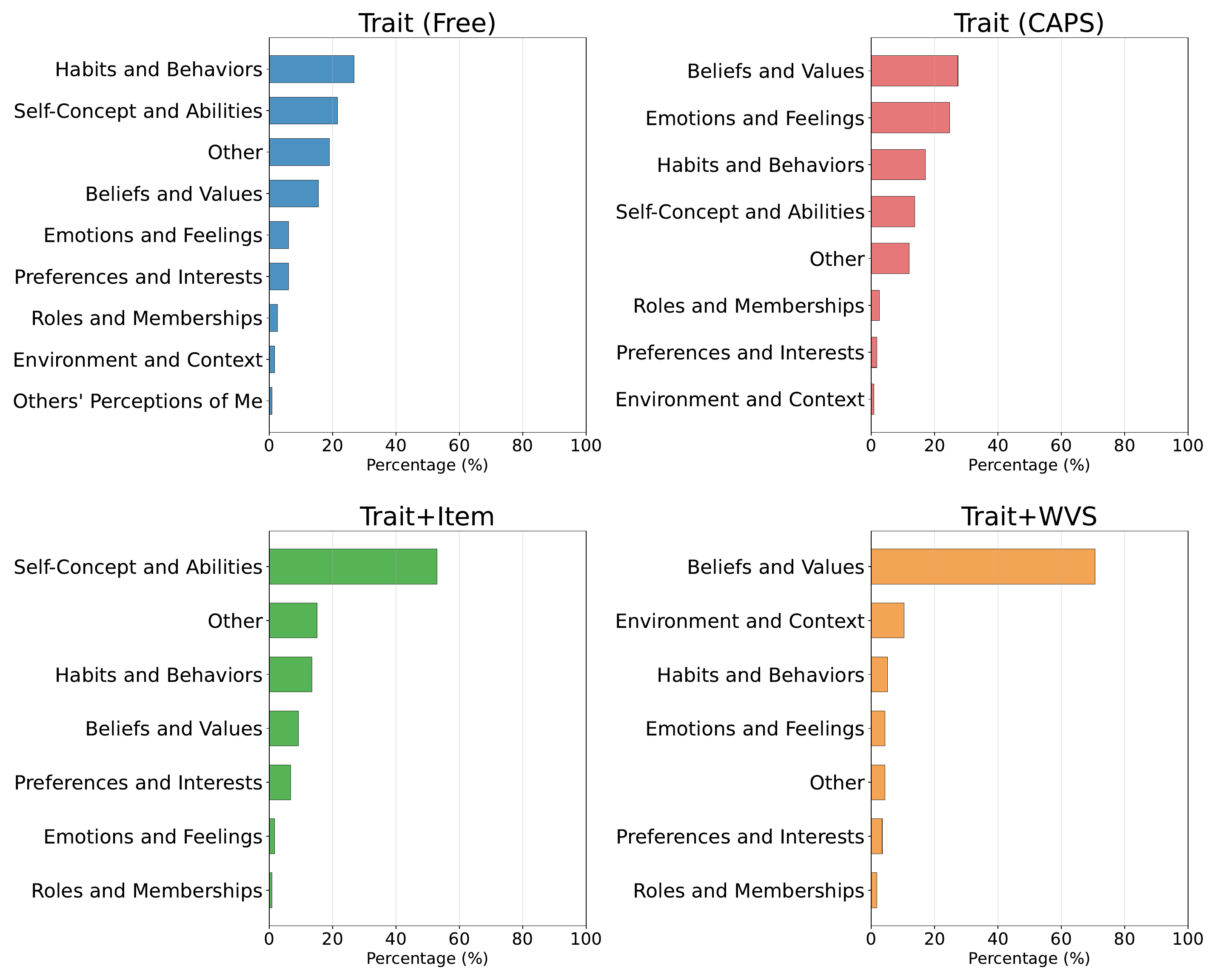}
    \caption{Mediator categories.}
    \label{fig:mediator_categoty}
\end{figure}

Figure~\ref{fig:mediator_categoty} shows that all strategies include mediators across multiple categories, with Trait (Free) and Trait (CAPS) displaying relatively balanced distributions.
Trait+Item has about half of its mediators in Self-Concept and Abilities, reflecting the self-descriptive nature of its reference items, while Trait+WVS shows a notable presence in Beliefs and Values, consistent with the nature of the WVS survey, which is largely composed of questions on human values.
Overall, these patterns indicate that the mediators do not merely contain superficial content but also capture diverse and in-depth aspects of human characteristics.

%%%%%%%%%% Section 5. Experiment Settings %%%%%%%%%%
\section{Experiment Settings}

\subsection{Evaluation Criteria}\label{sec:evaluation_criteria}
To evaluate the quality of each item, we consider (1) construct validity: how well an item measures the trait it is intended to measure, and (2) internal consistency reliability: the homogeneity of items that are intended to measure the same trait \cite{weiner2017handbook}.

To quantify construct validity, we need the \textbf{trait score} of each respondent for each trait $t$, calculated as the average of the respondent's answer scores to the corresponding official items:
\begin{equation}\label{eq:trait_score}
S_t \;=\; \frac{1}{N}\;\sum_{i=1}^N a_{i}
\end{equation}
where $N$ is the number of official items measuring trait $t$, and $a_i$ is the respondent's answer to item $i$.
For items designed to have a negative correlation with the target trait, the answer $a_i$ is inverted as $\bigl(L_{\max} + 1\bigr) - a_i$, where $L_{\max}$ denotes the maximum possible answer score.
Higher trait scores indicate that an individual possesses a greater level of the corresponding trait.

Construct validity is divided into convergent validity and discriminant validity as follows.

\paragraph{Convergent Validity.}\label{sec:convergent_validity}
Convergent validity (CV) refers to the degree to which an item correlates with other measures that assess the same target trait (i.e., official items in our case) \cite{surucu2020validity}. The CV score of each item $i$ is defined as the Spearman correlation between respondents' answers to that item and their trait scores:
\begin{equation}
\text{CV}_i = \mathrm{Spearman}\bigl(\{a_i^{(r)}\}_{r=1}^R, \{S_t^{(r)}\}_{r=1}^R \bigr)
\end{equation}
where $R$ is the number of respondents, $a_i^{(r)}$ is respondent $r$'s answer to item $i$, $S_t^{(r)}$ is respondent $r$'s trait score for the trait $t$ targeted by the item, and $\mathrm{Spearman}(\cdot)$ is the Spearman rank‐order correlation.
For negatively-correlated items, respondents' answers are inverted as in trait scores.

The CV score is used for two distinct purposes, which should be not confused. 
First, the CV score as a measure of an item's quality is computed based on the responses of \textit{human respondents}. The CV score of a selected item set is the average of these scores for the individual items in the set.
On the other hand, in the item selection stage of the simulation framework (\S\ref{sec:item_filtering}), generated items are ranked by their CV scores calculated from the responses of \textit{virtual respondents}.

We define two additional metrics based on the CV score to evaluate different aspects of mediator generation strategies.
First, to assess the relative quality of the selected item set, we compute the \textbf{percentile}, which represents the position of the selected item set’s CV score within the empirical distribution of CV scores derived from all possible combinations of $N$ items drawn from the initial item pool.
Second, to measure how accurately each mediator generation strategy ranks individual items, we compute the \textbf{normalized discounted cumulative gain (NDCG)}, using the ranking based on each item's ground-truth CV score as the reference. 
Ranking accuracy is particularly important for downscaling existing survey items.
To evaluate ranking accuracy across all items as well as within the selected $N$-item set, we report both NDCG and NDCG@$N$. The definitions of percentile and NDCG are detailed in Appendix~\ref{appendix:denfinition_percentile_ndcg}.

\paragraph{Discriminant Validity.} 
Discriminant validity (DV) refers to the extent to which an item exhibits low correlations with measures of different, theoretically unrelated traits \cite{surucu2020validity}.
The DV score of an item $i$ is defined as the mean of the Spearman correlations between respondents' answers to $i$ and their trait scores for \textit{non-target} traits:
\begin{equation}
\begin{split}
&\text{DV}_i =\\
&\frac{1}{|\mathcal{D}|}
\sum_{t \in \mathcal{D}} \Bigl|\mathrm{Spearman}\Bigl(\{a_i^{(r)}\}_{r=1}^R, \{S_t^{(r)}\}_{r=1}^R \Bigr)\Bigr|
\end{split}
\end{equation}
where $\mathcal{D}$ is the set of non-target traits, and the remaining notation follows that used for convergent validity.
We take the absolute values of the correlations because correlations in either direction (positive or negative) are undesirable.
The DV score of an item set is calculated as the average DV score of the items within the set.

While low DV scores are generally desirable, their interpretation warrants caution because survey items that are completely unrelated to psychometric assessment can still achieve scores close to 0.
Hence, we recommend interpreting DV scores in conjunction with CV scores (e.g., the ratio of DV to CV scores) to ensure that items are both non-divergent and meaningfully aligned with the intended traits.

\paragraph{Internal Consistency Reliability.}
Internal consistency reliability (ICR) refers to the extent to which a set of items homogeneously measures the same trait \cite{surucu2020validity}. We calculate the ICR score for a trait $t$ using Cronbach's alpha \cite{cronbach1951coefficient}, which is widely used in psychometrics:
\begin{equation}
\text{ICR}_t = \frac{N}{N-1}
\Biggl(1 - \frac{\sum_{i=1}^{N}\mathrm{Var}(\{a^{(r)}_i\}_{r=1}^{R})}{\mathrm{Var}\bigl(\{\sum_{i = 1}^N a^{(r)}_i\}_{r=1}^{R} \bigr)}\Biggr)
\end{equation}
where $N$ is the number of items measuring each trait $t$, and $a^{(r)}_i$ is respondent $r$'s answer to item $i$.
Higher ICR scores indicate that the set of items homogeneously measures a single trait.
A Cronbach's alpha above 0.7 is generally considered acceptable \cite{surucu2020validity}.

\subsection{Baselines} 
As baseline methods for item selection, we employ three approaches: (1) random ordering, (2) LLM-as-a-judge, and (3) no-mediator.
The prompts for (1) and (2) are provided in Appendix~\ref{appendix:baseline}.

\paragraph{Random Ordering.}
Items are ranked randomly, and the highest-ranked items are selected.

\paragraph{LLM-as-a-Judge.}
We use GPT-4.1-mini to directly evaluate the quality of each item based on three criteria: convergent validity, discriminant validity (\S\ref{sec:evaluation_criteria}), and test-retest reliability.
We cannot use internal consistency reliability here because it is not defined at the item level.
Instead, we include test-retest reliability, which measures the consistency of responses when the same respondent completes the same item at two different time points \citep{surucu2020validity}.

We provide definitions for each criterion, and the LLM predicts numerical scores ranging from 1 to 100 for each.
The final item quality is quantified as the average score across the three criteria, computed over 10 runs to account for variability in LLM outputs.

\paragraph{No-Mediator.}
To examine the importance of the mediator, we removed the mediator from the simulation prompt, allowing us to compare performance with and without its presence.

\subsection{Experimental Setup}
Mediator generation is performed using GPT-4.1, while both the mediator-guided simulation and the LLM-as-a-judge baseline use GPT-4.1-mini.
For simulations, we use 500 virtual respondents with distinct persona profiles by default, and only the mediator parts vary across mediator generation strategies. 
We employ GPT-4.1-mini as the simulation model and set the temperature to 0 to ensure better reproducibility.
For each item selection method, we select the top $N$ items for each trait, where $N$ matches the number of items in the official survey (i.e., 10 for Big5 and 4 for Schwartz and VIA).

%%%%%%%%%% Section 6. Results %%%%%%%%%%
\section{Results}\label{sec:results}

% Table: Main Results
\begin{table}[!t]\footnotesize
\begin{center}
\begin{tabular}{l@{\hskip 8pt}c@{\hskip 8pt}c@{\hskip 6pt}c@{\hskip 6pt}c@{\hskip 8pt}c@{\hskip 8pt}c}
\toprule
    \textbf{Method} & \multicolumn{4}{c}{\textbf{CV$\uparrow$}} & \textbf{DV$\downarrow$} & \textbf{ICR$\uparrow$} \\
    \cmidrule{2-7}
     & \scriptsize Score & \scriptsize Per. & \scriptsize NDCG & \scriptsize @$N$ & \scriptsize Score & \scriptsize Score \\
\midrule
    \multicolumn{7}{c}{Big5} \\
\midrule
    Random & .546 & 48.4 & .374 & .140 & .290 & .848 \\
    LLM-Judge & .599 & 79.7 & .361 & .166 & \underline{.287} & \underline{.897} \\
    No-Mediator & .585 & 82.2 & .441 & .269 & .298 & .895 \\
    Trait (Free) & \textbf{.632} & \textbf{99.3} & \textbf{.568} & \underline{.455} & .294 & \textbf{.904}\\
    Trait (CAPS) & .587 & 71.0 & .521 & .333 & .296 & .892 \\
    Trait+Item & \underline{.626} & \underline{98.3} & \underline{.552} & \textbf{.473} & .294 & .892 \\
    Trait+WVS & .614 & 95.5 & .429 & .282 & .302 & \textbf{.904} \\
    Sampling & .516 & 30.6 & .380 & .149 & \textbf{.265} & .839 \\
\midrule
    Oracle & .690 & 100 & 1 & 1 & .318 & .922 \\
    Official & .657 & - & - & - & .271 & .844 \\
\midrule
    \multicolumn{7}{c}{Schwartz's Theory of Basic Values} \\
\midrule
    Random & .258 & 61.1 & .525 & .256 & .140 & .576 \\
    LLM-Judge & .332 & \underline{86.8} & .642 & \underline{.454} & .143 & .679 \\
    No-Mediator & \underline{.333} & 86.3 & .605 & .397 & .139 & \underline{.756} \\
    Trait (Free) & .313 & 77.9 & \underline{.661} & .439 & .143 & \textbf{.757} \\
    Trait (CAPS) & \textbf{.347} & \textbf{87.1} & \textbf{.664} & \textbf{.492} & .145 & .740 \\
    Trait+Item & .327 & 83.1 & .630 & .427 & .140 & .699 \\
    Trait+WVS & \underline{.333} & 85.4 & .645 & .419 & \textbf{.133} & .748 \\
    Sampling & .304 & 75.6 & .531 & .284 & \underline{.137} & .684 \\
\midrule
    Oracle & .432 & 100 & 1 & 1 & .146 & .724 \\
    Official & .605 & - & - & - & .169 & .711 \\
\midrule
    \multicolumn{7}{c}{VIA} \\
\midrule
    Random & .492 & 48.0 & .504 & .223 & .293 & .683 \\
    LLM-Judge & .561 & 75.5 & .542 & .275 & .292 & \underline{.790} \\
    No-Mediator & .502 & 47.9 & .482 & .206 & .281 & .717 \\
    Trait (Free) & \textbf{.586} & \textbf{88.5} & \underline{.657} & \underline{.456} & .299 & \textbf{.803} \\
    Trait (CAPS) & .557 & 75.1 & .573 & .345 & .296 & .772 \\
    Trait+Item & \underline{.573} & \underline{78.2} & \textbf{.678} & \textbf{.500} & \textbf{.276} & .753 \\
    Trait+WVS & .529 & 61.4 & .508 & .244 & .299 & .745 \\
    Sampling & .528 & 63.0 & .562 & .286 & \underline{.286} & .710 \\
\midrule
    Oracle & .658 & 100 & 1 & 1 & .300 & .837 \\
    Official & .765 & - & - & - & .278 & .760 \\
\bottomrule
\end{tabular}
\end{center}
\caption{Performance of item selection methods. Metrics represent the average performance of item sets across all traits within each survey. \textbf{Bold}: best performance, \underline{Underlined}: second-best performance. (Per.: percentile, NDCG: NDCG@All, @$N$: NDCG@$N$).}
\label{tab:result_table}
\end{table}

Table~\ref{tab:result_table} presents the overall performance of different methods.
As a reference point, we report the performance of the \textbf{oracle item set}, i.e., the set of $N$ generated items yielding the highest CV score based on human responses.
For meaningful comparisons between psychologist-authored and LLM-generated items, we also present the performance of the \textbf{official items} as if they had been generated and selected as the final item set.
The significance test results are in Appendix~\ref{appendix:significance_test_main}, where the p-values are used to characterize the stability of performance differences across repeated stochastic runs.

\subsection{Convergent Validity}
Results demonstrate that the trait-only generation strategies are most effective for identifying items with high convergent validity across the three surveys.
The free generation strategy (Trait (Free)) achieved the best CV score and percentile on the Big5 and VIA, while the CAPS-guided strategy (Trait (CAPS)) yielded the best performance on the Schwartz. NDCG and NDCG@$N$ scores show similar patterns.
Overall, the quality of the selected item sets ranks in the top 1\% (Big5) to 13\% (Schwartz and VIA) among all possible selections, according to the percentile metric. 

Our simulation-based approach, when paired with the best-performing mediators, consistently outperforms baseline methods.
The random ordering approach produced results close to random, yielding approximately the 50th percentile.
The no-mediator approach consistently underperformed compared to the best mediator-based methods, underscoring the critical role of mediators in item validation.
Notably, the LLM-as-a-judge approach consistently scored above the 75th percentile across all surveys, indicating that LLMs possess a solid understanding of psychological traits and a reasonable ability to evaluate item quality.
However, this method underperforms in ranking individual items, as indicated by lower NDCG scores compared to simulation-based methods.

Finally, using real human demographics as mediators yielded poorer performance than using LLM-generated mediators.
This finding aligns with the prior finding that LLMs struggle to capture the relationship between demographic variables and psychological traits \cite{petrov2024limited}.

These results suggest that LLMs have the ability to generate effective mediators between traits and behaviors using only trait names and definitions, without additional context.
Still, a significant performance gap remains between all methods and the oracle and official item sets. 

\subsection{Discriminant Validity}
Discriminant validity (DV) measures an item's correlation with unrelated traits.
The results show that all methods yielded DV scores similar to the oracle and official items, suggesting acceptable levels of DV.
While simulation-based methods and baselines show marginal differences, the ratio of DV to CV scores---interpreted as the relative correlation with non-target traits compared to target traits---favors the Trait (Free) strategy.
Overall, these findings demonstrate LLMs' ability to reasonably generate items without unintended correlations with unrelated traits.
In Appendix~\ref{appendix:dv_max}, we report the maximum DV scores for each survey to examine worst cases.

\subsection{Internal Consistency Reliability}\label{sec:ICR}
Internal consistency reliability (ICR) evaluates whether items within a set capture the same trait.
Trait (Free) consistently achieved the highest ICR scores across all surveys.
Furthermore, most item sets selected through our mediator-driven approach achieved ICR scores above the acceptance threshold ($\ge$ 0.7).
These results reaffirm the utility of LLM-generated mediators and the effectiveness of mediator-guided simulation in capturing the homogeneity of items.
ICR scores have a positive association with CV scores, suggesting that securing high convergent validity enhances the homogeneity of the items.

%%%%%%%%%% Section 7. Ablation Studies %%%%%%%%%%
\section{Ablation Studies}\label{sec:ablation_study}
In this section, we conduct ablation studies to examine the importance of each component of the simulation.
We use the best-performing mediator generation strategy for each survey identified in Section~\ref{sec:results}, namely Trait (Free) for Big5 and VIA, and Trait (CAPS) for Schwartz.

\subsection{Simulation Components}\label{sec:simulation_components}

% Figure: Simulation Components
\begin{figure}[!t]
    \centering
    \includegraphics[width=0.48\textwidth]{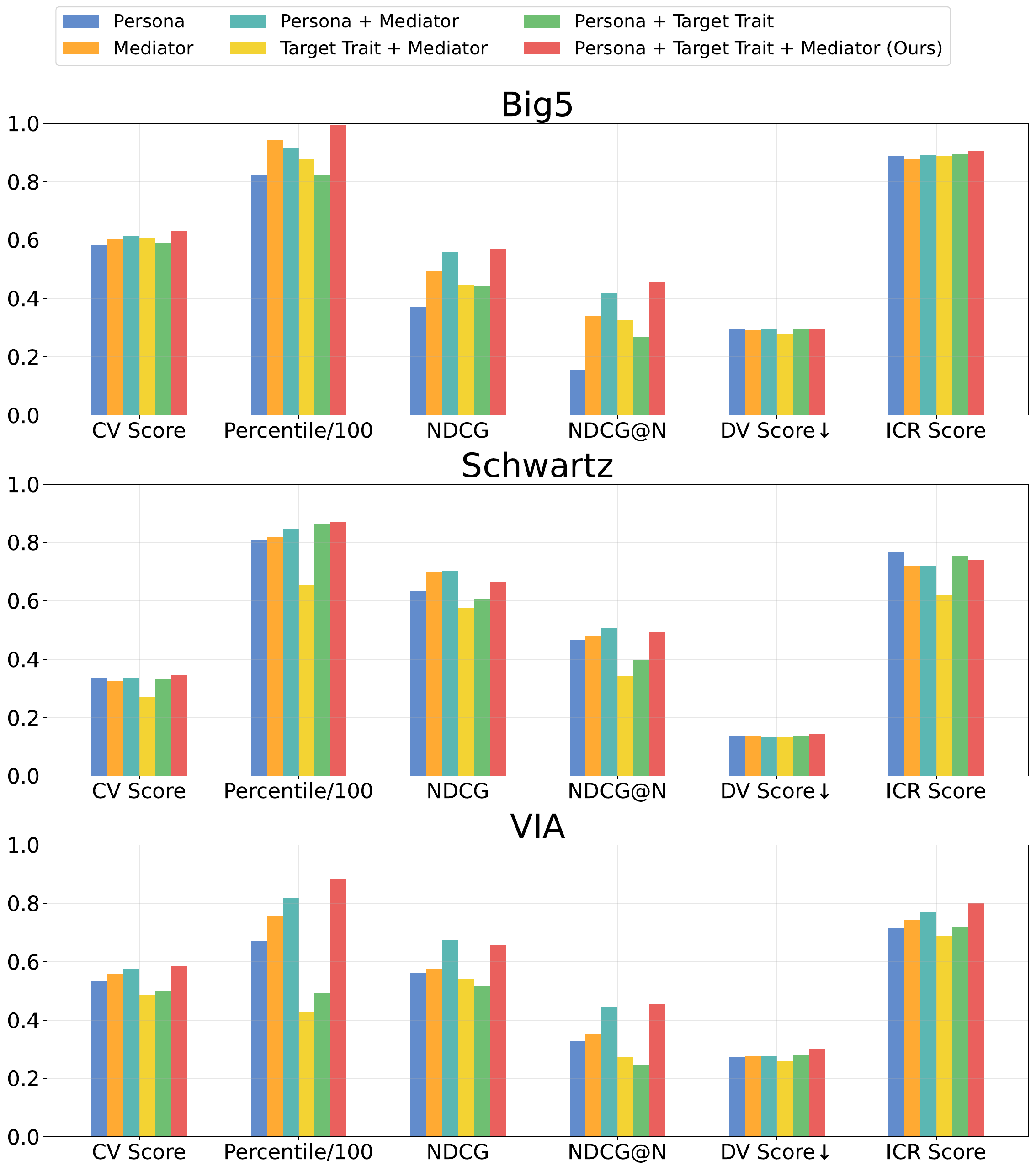}
    \caption{Performance of selected items across different components of the simulation prompt. Percentile scores were normalized to the [0, 1] range.}
    \label{fig:simulation_component}
\end{figure}

Aside from the essential task instructions with survey items and answer choices, our simulation prompt comprises three core components: the target trait, mediator, and general persona profile.
To examine the importance of each component, we evaluate item selection performance across all possible combinations. Since the persona profile serves to distinguish among 500 virtual respondents, when it is absent, we generate distinct respondents based on the number of mediators. We do not include using a target trait only, because a single target trait sentence cannot generate distinct respondents. In total, we evaluate six combinations.
The significance test results are in Appendix~\ref{appendix:significance_test_ablation}.

Figure~\ref{fig:simulation_component} shows that the full setting---combining sentences that steer target traits and mediators with persona profiles---most effectively identifies valid item sets.
This setting achieves the highest CV scores and percentile across all three surveys, as well as the highest ICR scores in both Big5 and VIA.
When comparing Persona with Persona+Mediator and Persona+Target Trait with the full setting, we observe that removing mediators leads to the largest performance drop in CV, percentile, and ranking scores. Furthermore, comparing Target Trait+Mediator with the full setting also reveals declines across all metrics. These findings suggest that both mediators and persona profiles play a crucial role in identifying valid survey items.

\subsection{Simulation Scale}\label{sec:simulation_scale}
We examine the relationship between simulation scale and item selection performance by varying the number of virtual respondents from 50 to 500.
For scales below 500, we sample persona profiles from the full set of 500. To establish confidence interval, we perform subsampling without replacement 1,000 times.

% Figure: Simulation Scale
\begin{figure}[!t]
    \centering
    \includegraphics[width=0.492\textwidth]{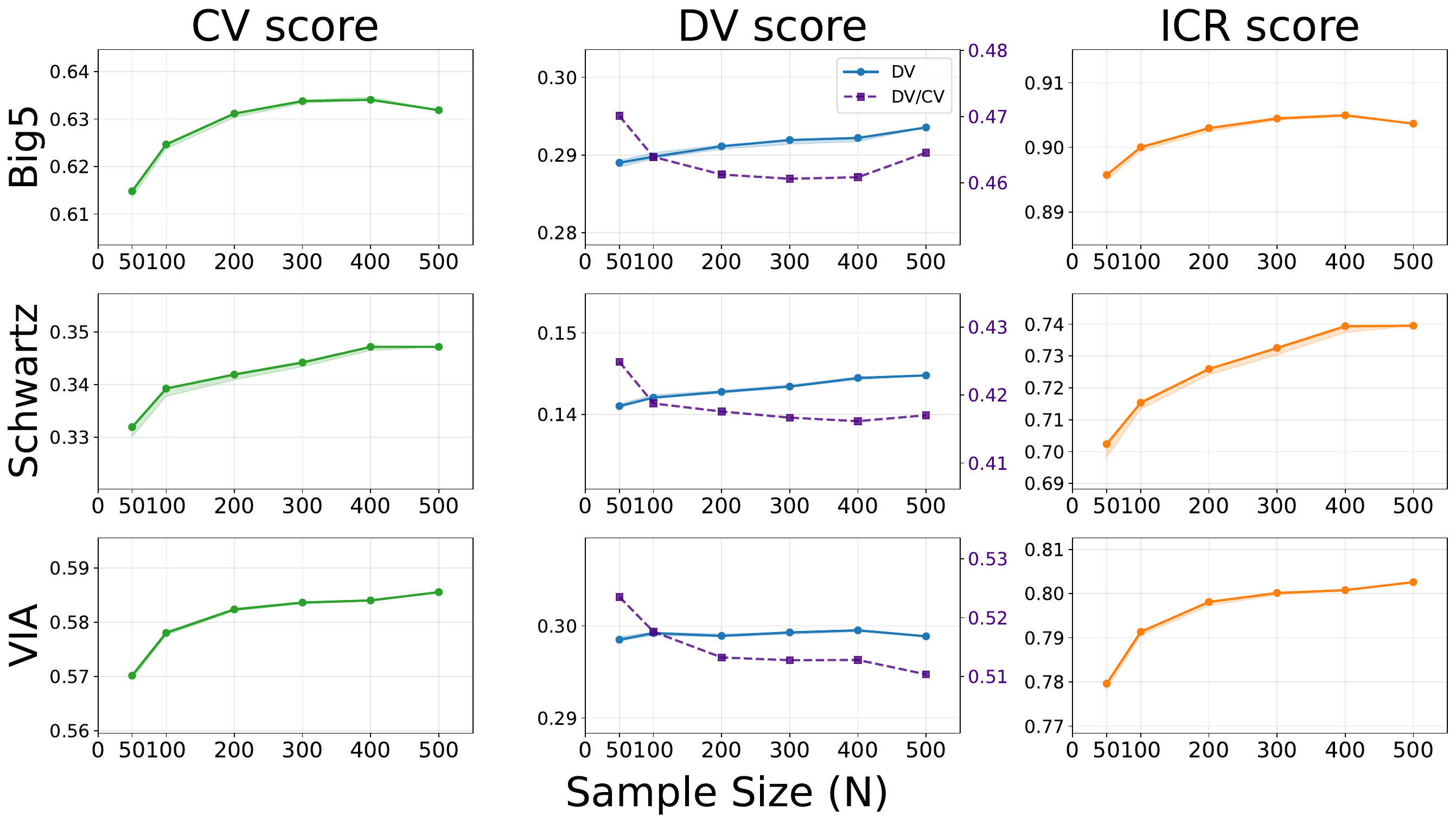}
    \caption{Performance by number of virtual respondents. Points indicate median scores, with shaded regions representing confidence intervals.}
    \label{fig:simulation_scale}
\end{figure}

Figure~\ref{fig:simulation_scale} shows that increasing the simulation scale has a positive impact on CV and ICR scores, with an upward trend. This trend is pronounced for ICR scores. 
DV scores remain relatively stable (VIA) or show a slight upward trend (Big5 and Schwartz).
However, DV scores should be considered in conjunction with CV scores, since items with low correlations with any traits can achieve low DV scores.
The figure shows that the ratio of DV to CV scores decreases, suggesting that larger scales make it easier to identify items that are more closely aligned with their target traits, as CV grows faster than DV.
The results of this scaling experiment align with real-world practice, where securing a large pool of human respondents is desirable for both the validity and reliability of survey items.

\subsection{Simulation Models}\label{sec:simulation_models}

To assess the generalizability of our framework across different LLMs serving as simulation models, we conduct simulations with three additional models alongside GPT-4.1-mini: GPT-4.1-nano, LLaMA-4-Scout, and LLaMA-3.1-70B.

\begin{figure}[!t]
    \centering
    \includegraphics[width=0.49\textwidth]{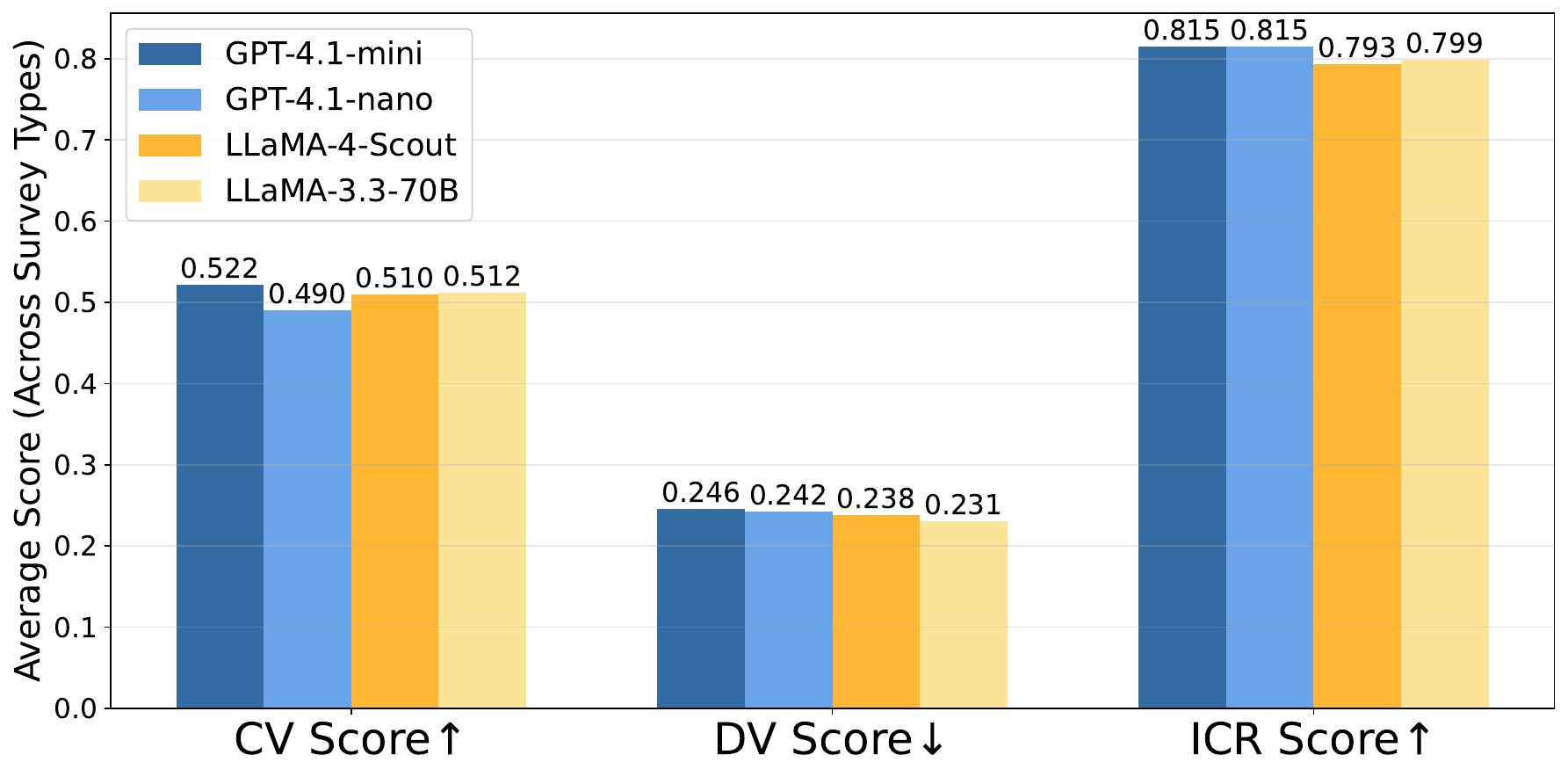}
    \caption{Performance by simulation LLM (with 500 virtual respondents).}
    \label{fig:simulation_model}
\end{figure}

Figure~\ref{fig:simulation_model} shows that our framework achieves consistent performance across different simulation models, with only marginal differences observed among the four.
Notably, the performance of GPT-series models is positively correlated with their sizes, whereas LLaMA-series models show smaller performance differences across sizes.

%%%%%%%%%% Section 8. Conclusion %%%%%%%%%%
\section{Conclusion}
With the goal of evaluating the construct validity of psychometric survey items, we presented a framework for the automatic generation and validation of survey items.
We focused particularly on mediator generation using LLMs and mediator-guided simulation with LLM-based virtual respondents. 
We also defined psychometrically grounded metrics for this new task.
Experiments demonstrated that our framework effectively identifies valid item sets.
In particular, mediators generated by LLMs based on trait definitions performed best overall, highlighting the ability of LLMs to generate mediators and simulate survey respondents.

To support follow-up research, we publicly release all generated items along with collected human responses.
We hope this benchmark serves as a valuable resource for research on improved methods for more scalable and cheaper survey item validation.

%%%%%%%%%% Section 9. Limitations %%%%%%%%%%
\section{Limitations}
We focused on three psychological theories, which may not capture the full breadth of psychological domains such as emotion, cognition, and mental disorders. However, our framework can be readily extended to other theories, which we leave for future work.

The generated items are currently only in English, and our framework has not been validated in multilingual or cross-cultural settings.
Future work may extend the method by generating psychometric items in other languages and leveraging multilingual versions of official items (which already exist for many theories). The item ranking results can be validated by comparing them with human responses collected in that language or cultural background.

Our framework assumes the availability of official items to compute validity.
While this is a common assumption in real-world survey refinement, it would be valuable to explore item evaluation from scratch.
For example, factor analysis is a statistical method widely used to uncover latent structures underlying responses and identify groups of items loading on the same factor.
Our simulation method can provide virtual survey response data to support factor analysis and thereby facilitate the development of new items in domains where official items do not yet exist or are not accessible.

As this study represents an early step toward automated item validation and there remains substantial room for improvement, we recommend conducting a human inspection before officially releasing new psychometric items developed using our framework. Experts may verify that each item measures the intended trait and is free of conceptual or interpretive errors. Our framework can be integrated at the initial stage of the validation process, where it narrows a large candidate pool to a smaller set of high-potential items.

It is important to note that LLMs cannot yet perfectly replicate the responses of individual humans.
Our framework is therefore not intended to mimic human psychological processes per se, but to leverage diverse mediators and their correlations with trait scores to identify items that measure traits robustly.
Recognizing this inherent limitation of LLMs, we view our work as a meaningful step toward more effective automatic item validation, and encourage future research to advance simulation techniques that better align with human psychological processes.

Additionally, we note that LLMs do not yet possess a perfect ability to generate mediators.
In our analysis of failure cases, some mediators described characteristics of individuals already high in the target trait, rather than interacting with the trait as intended.
This suggests the need for future work on improved prompting strategies and automated methods for filtering generated mediators, in line with our goal of fully automatic item validation.

\section*{Acknowledgments}
This work was supported by the National Research Foundation of Korea (NRF) (RS-2024-00333484, RS-2022-NR070855) and by the Institute of Information \& Communications Technology Planning \& Evaluation (IITP) (RS-2021-II211343, AI Graduate School Program, Seoul National University), all funded by the Korean government (MSIT).
This work was also supported by the Creative-Pioneering Researchers Program through Seoul National University and by the National Supercomputing Center with supercomputing resources including technical support (KSC-2025-CRE-0332, KSC-2025-CRE-0514).

\bibliography{main}

@inproceedings{miotto-etal-2022-gpt,
    title = "Who is {GPT}-3? {A}n exploration of personality, values and demographics",
    author = "Miotto, Maril{\`u}  and
      Rossberg, Nicola  and
      Kleinberg, Bennett",
    editor = "Bamman, David  and
      Hovy, Dirk  and
      Jurgens, David  and
      Keith, Katherine  and
      O'Connor, Brendan  and
      Volkova, Svitlana",
    booktitle = "Proceedings of the Fifth Workshop on Natural Language Processing and Computational Social Science (NLP+CSS)",
    month = nov,
    year = "2022",
    address = "Abu Dhabi, UAE",
    publisher = "Association for Computational Linguistics",
    url = "https://aclanthology.org/2022.nlpcss-1.24/",
    doi = "10.18653/v1/2022.nlpcss-1.24",
    pages = "218--227",
}

@article{ye2025large,
  title={{Large Language Model Psychometrics: A Systematic Review of Evaluation, Validation, and Enhancement}},
  author={Ye, Haoran and Jin, Jing and Xie, Yuhang and Zhang, Xin and Song, Guojie},
  journal={arXiv preprint arXiv:2505.08245},
  year={2025},
  url={https://arxiv.org/abs/2505.08245}
}

@inproceedings{huang-etal-2024-reliability,
    title = "{On the Reliability of Psychological Scales on Large Language Models}",
    author = "Huang, Jen-tse  and
      Jiao, Wenxiang  and
      Lam, Man Ho  and
      Li, Eric John  and
      Wang, Wenxuan  and
      Lyu, Michael",
    editor = "Al-Onaizan, Yaser  and
      Bansal, Mohit  and
      Chen, Yun-Nung",
    booktitle = "Proceedings of the 2024 Conference on Empirical Methods in Natural Language Processing",
    month = nov,
    year = "2024",
    address = "Miami, Florida, USA",
    publisher = "Association for Computational Linguistics",
    url = "https://aclanthology.org/2024.emnlp-main.354/",
    doi = "10.18653/v1/2024.emnlp-main.354",
    pages = "6152--6173",
}

@inproceedings{lee-etal-2025-llms,
    title = "Do {LLM}s Have Distinct and Consistent Personality? {TRAIT}: Personality Testset designed for {LLM}s with Psychometrics",
    author = "Lee, Seungbeen  and
      Lim, Seungwon  and
      Han, Seungju  and
      Oh, Giyeong  and
      Chae, Hyungjoo  and
      Chung, Jiwan  and
      Kim, Minju  and
      Kwak, Beong-woo  and
      Lee, Yeonsoo  and
      Lee, Dongha  and
      Yeo, Jinyoung  and
      Yu, Youngjae",
    editor = "Chiruzzo, Luis  and
      Ritter, Alan  and
      Wang, Lu",
    booktitle = "Findings of the Association for Computational Linguistics: NAACL 2025",
    month = apr,
    year = "2025",
    address = "Albuquerque, New Mexico",
    publisher = "Association for Computational Linguistics",
    url = "https://aclanthology.org/2025.findings-naacl.469/",
    doi = "10.18653/v1/2025.findings-naacl.469",
    pages = "8412--8452",
    ISBN = "979-8-89176-195-7"
}

@article{lindeman,
author = {Lindeman, Marjaana and Verkasalo, Markku},
year = {2005},
month = {11},
pages = {170-8},
title = {{Measuring Values With the Short Schwartz's Value Survey}},
volume = {85},
journal = {Journal of personality assessment},
doi = {10.1207/s15327752jpa8502_09}
}

@article{RAMMSTEDT2007203,
title = {{Measuring personality in one minute or less: A 10-item short version of the Big Five Inventory in English and German}},
journal = {Journal of Research in Personality},
volume = {41},
number = {1},
pages = {203-212},
year = {2007},
issn = {0092-6566},
doi = {https://doi.org/10.1016/j.jrp.2006.02.001},
url = {https://www.sciencedirect.com/science/article/pii/S0092656606000195},
author = {Beatrice Rammstedt and Oliver P. John}
}

@article{neo-pi-r,
author = {Costa, Paul and McCrae, Robert},
year = {2008},
month = {01},
pages = {179-198},
title = {{The Revised NEO Personality Inventory (NEO-PI-R)}},
volume = {2},
journal = {The SAGE Handbook of Personality Theory and Assessment},
doi = {10.4135/9781849200479.n9}
}

@article{surucu2020validity,
author = {Sürücü, Lütfi and Maslakci, Ahmet},
year = {2020},
month = {10},
pages = {2694-2726},
title = {{Validity And Reliability In Quantitative Research}},
volume = {8},
journal = {Business And Management Studies An International Journal},
doi = {10.15295/bmij.v8i3.1540}
}

@article{weber2002domain,
author = {Weber, Elke U. and Blais, Ann-Renée and Betz, Nancy E.},
title = {A domain-specific risk-attitude scale: {M}easuring risk perceptions and risk behaviors},
journal = {Journal of Behavioral Decision Making},
volume = {15},
number = {4},
pages = {263-290},
doi = {https://doi.org/10.1002/bdm.414},
url = {https://onlinelibrary.wiley.com/doi/abs/10.1002/bdm.414},
eprint = {https://onlinelibrary.wiley.com/doi/pdf/10.1002/bdm.414},
year = {2002}
}

@article{mccrae1997personality,
  title={Personality trait structure as a human universal.},
  author={McCrae, Robert R. and Costa Jr, Paul T.},
  journal={American psychologist},
  volume={52},
  number={5},
  pages={509},
  year={1997},
  publisher={American Psychological Association},
  doi={https://doi.org/10.1037/0003-066X.52.5.509}
}

@article{mischel1995cognitive,
  title={A cognitive-affective system theory of personality: {R}econceptualizing situations, dispositions, dynamics, and invariance in personality structure.},
  author={Mischel, Walter and Shoda, Yuichi},
  journal={Psychological review},
  volume={102},
  number={2},
  pages={246},
  year={1995},
  publisher={American Psychological Association},
  doi={https://doi.org/10.1037/0033-295X.102.2.246}
}

@inproceedings{jiang2023evaluating,
 author = {Jiang, Guangyuan and Xu, Manjie and Zhu, Song-Chun and Han, Wenjuan and Zhang, Chi and Zhu, Yixin},
 booktitle = {Advances in Neural Information Processing Systems},
 editor = {A. Oh and T. Naumann and A. Globerson and K. Saenko and M. Hardt and S. Levine},
 pages = {10622--10643},
 publisher = {Curran Associates, Inc.},
 title = {{Evaluating and Inducing Personality in Pre-trained Language Models}},
 url = {https://proceedings.neurips.cc/paper_files/paper/2023/file/21f7b745f73ce0d1f9bcea7f40b1388e-Paper-Conference.pdf},
 volume = {36},
 year = {2023}
}

@inproceedings{han-etal-2025-value,
    title = "{Value Portrait: Assessing Language Models' Values through Psychometrically and Ecologically Valid Items}",
    author = "Han, Jongwook  and
      Choi, Dongmin  and
      Song, Woojung  and
      Lee, Eun-Ju  and
      Jo, Yohan",
    editor = "Che, Wanxiang  and
      Nabende, Joyce  and
      Shutova, Ekaterina  and
      Pilehvar, Mohammad Taher",
    booktitle = "Proceedings of the 63rd Annual Meeting of the Association for Computational Linguistics (Volume 1: Long Papers)",
    month = jul,
    year = "2025",
    address = "Vienna, Austria",
    publisher = "Association for Computational Linguistics",
    url = "https://aclanthology.org/2025.acl-long.838/",
    doi = "10.18653/v1/2025.acl-long.838",
    pages = "17119--17159",
    ISBN = "979-8-89176-251-0"
}

@article{Hadar_Shoval,
author = {Hadar Shoval, Dorith and Asraf, Kfir and Mizrachi, Yonathan and Haber, Yuval and Elyoseph, Zohar},
year = {2024},
month = {04},
pages = {},
title = {{Assessing the Alignment of Large Language Models With Human Values for Mental Health Integration: Cross-Sectional Study Using Schwartz’s Theory of Basic Values}},
journal = {JMIR Mental Health},
doi = {10.2196/55988}
}

@article{goldberg1999broad,
  title={A broad-bandwidth, public domain, personality inventory measuring the lower-level facets of several {F}ive-{F}actor models},
  author={Goldberg, Lewis R.},
  journal={Personality psychology in Europe},
  volume={7},
  number={1},
  pages={7--28},
  year={1999},
  publisher={Tilburg Netherland}
}

@book{jung2016psychological,
  title={Psychological {T}ypes},
  author={Jung, Carl and Beebe, John},
  year={2016},
  publisher={Routledge},
  doi={https://doi.org/10.4324/9781315512334}
}

@article{beck1996comparison,
  title={{Comparison of Beck Depression Inventories-IA and-II in Psychiatric Outpatients}},
  author={Beck, Aaron T. and Steer, Robert A. and Ball, Roberta and Ranieri, William F.},
  journal={Journal of personality assessment},
  volume={67},
  number={3},
  pages={588--597},
  year={1996},
  publisher={Taylor \& Francis},
  doi={10.1207/s15327752jpa6703_13}
}

@article{beck1988inventory,
  title={An inventory for measuring clinical anxiety: {P}sychometric properties.},
  author={Beck, Aaron T. and Epstein, Norman and Brown, Gary and Steer, Robert A.},
  journal={Journal of consulting and clinical psychology},
  volume={56},
  number={6},
  pages={893},
  year={1988},
  publisher={American Psychological Association},
  doi={10.1037//0022-006x.56.6.893}
}

@article{mulla2023automatic,
  title={Automatic question generation: {A} review of methodologies, datasets, evaluation metrics, and applications},
  author={Mulla, Nikahat and Gharpure, Prachi},
  journal={Progress in Artificial Intelligence},
  volume={12},
  number={1},
  pages={1--32},
  year={2023},
  publisher={Springer},
  doi={10.1007/s13748-023-00295-9}
}

@inproceedings{laverghetta2024creative,
  title={The creative psychometric item generator: {A} framework for item generation and validation using large language models},
  author={Laverghetta, Antonio and Luchini, Simone and Linnell, Averie and Reiter-Palmon, Roni and Beaty, Roger},
  booktitle={CEUR Workshop Proceedings},
  volume={3810},
  pages={59--73},
  year={2024},
  organization={CEUR-WS}
}

@ARTICLE{10522667,
  author={Babakhani, Pedram and Lommatzsch, Andreas and Brodt, Torben and Sacker, Doreen and Sivrikaya, Fikret and Albayrak, Sahin},
  journal={IEEE Access}, 
  title={{Opinerium: Subjective Question Generation Using Large Language Models}}, 
  year={2024},
  volume={12},
  number={},
  pages={66085-66099},
  doi={10.1109/ACCESS.2024.3398553}
}

@article{golafshani2003understanding,
  title={{Understanding Reliability and Validity in Qualitative Research}},
  author={Golafshani, Nahid},
  journal={The qualitative report},
  volume={8},
  number={4},
  pages={597--607},
  year={2003},
  publisher={Canad{\'a}},
  doi={10.46743/2160-3715/2003.1870}
}

@article{argyle2023out,
  title={{Out of One, Many: Using Language Models to Simulate Human Samples}},
  volume={31},
  DOI={10.1017/pan.2023.2},
  number={3},
  journal={Political Analysis},
  author={Argyle, Lisa P. and Busby, Ethan C. and Fulda, Nancy and Gubler, Joshua R. and Rytting, Christopher and Wingate, David},
  year={2023},
  pages={337–351}
}

@article{liu2025leveraging,
  title={{Leveraging LLM respondents for item evaluation: A psychometric analysis}},
  author={Liu, Yunting and Bhandari, Shreya and Pardos, Zachary A.},
  journal={British Journal of Educational Technology},
  year={2025},
  publisher={Wiley Online Library},
  doi={https://doi.org/10.1111/bjet.13570}
}

@InProceedings{aher2023using,
  title = 	 {{Using Large Language Models to Simulate Multiple Humans and Replicate Human Subject Studies}},
  author =       {Aher, Gati V. and Arriaga, Rosa I. and Kalai, Adam Tauman},
  booktitle = 	 {Proceedings of the 40th International Conference on Machine Learning},
  pages = 	 {337--371},
  year = 	 {2023},
  editor = 	 {Krause, Andreas and Brunskill, Emma and Cho, Kyunghyun and Engelhardt, Barbara and Sabato, Sivan and Scarlett, Jonathan},
  volume = 	 {202},
  series = 	 {Proceedings of Machine Learning Research},
  month = 	 {23--29 Jul},
  publisher =    {PMLR},
  url = 	 {https://proceedings.mlr.press/v202/aher23a.html}
}

@article{petrov2024limited,
  title={{Limited Ability of LLMs to Simulate Human Psychological Behaviours: {A} Psychometric Analysis}},
  author={Petrov, Nikolay B. and Serapio-Garc{\'\i}a, Gregory and Rentfrow, Jason},
  journal={arXiv preprint arXiv:2405.07248},
  year={2024},
  url={https://arxiv.org/abs/2405.07248}
}

@article{goldberg1992development,
  title={The development of markers for the {B}ig-{F}ive factor structure.},
  author={Goldberg, Lewis R.},
  journal={Psychological assessment},
  volume={4},
  number={1},
  pages={26},
  year={1992},
  publisher={American Psychological Association},
  doi={https://doi.org/10.1037/1040-3590.4.1.26}
}

@incollection{schwartz1992universals,
title = {{Universals in the Content and Structure of Values: Theoretical Advances and Empirical Tests in 20 Countries}},
editor = {Mark P. Zanna},
series = {Advances in Experimental Social Psychology},
publisher = {Academic Press},
volume = {25},
pages = {1-65},
year = {1992},
issn = {0065-2601},
doi = {https://doi.org/10.1016/S0065-2601(08)60281-6},
url = {https://www.sciencedirect.com/science/article/pii/S0065260108602816},
author = {Shalom H. Schwartz}
}

@book{peterson2004character,
  title={{Character Strengths and Virtues: A Handbook and Classification}},
  author={Peterson, Christopher},
  volume={3},
  year={2004},
  publisher={Oxford University Press}
}

@article{yao2025clave,
  title={{CLAVE: An Adaptive Framework for Evaluating Values of LLM Generated Responses}},
  author={Yao, Jing and Yi, Xiaoyuan and Xie, Xing},
  journal={Advances in Neural Information Processing Systems},
  volume={37},
  pages={58868--58900},
  year={2025}
}

@inproceedings{zhang-etal-2018-personalizing,
    title = "{Personalizing Dialogue Agents: I have a dog, do you have pets too?}",
    author = "Zhang, Saizheng  and
      Dinan, Emily  and
      Urbanek, Jack  and
      Szlam, Arthur  and
      Kiela, Douwe  and
      Weston, Jason",
    editor = "Gurevych, Iryna  and
      Miyao, Yusuke",
    booktitle = "Proceedings of the 56th Annual Meeting of the Association for Computational Linguistics (Volume 1: Long Papers)",
    month = jul,
    year = "2018",
    address = "Melbourne, Australia",
    publisher = "Association for Computational Linguistics",
    url = "https://aclanthology.org/P18-1205/",
    doi = "10.18653/v1/P18-1205",
    pages = "2204--2213",
}

@book{weiner2017handbook,
  title={{Handbook of Personality Assessment}},
  author={Weiner, Irving B. and Greene, Roger L.},
  year={2017},
  publisher={John Wiley \& Sons}
}

@article{cronbach1951coefficient,
  title={{Coefficient Alpha and the Internal Structure of Tests}},
  volume={16},
  DOI={10.1007/BF02310555},
  number={3},
  journal={Psychometrika},
  author={Cronbach, Lee J.},
  year={1951},
  pages={297–334}
}

@article{goldberg1990bigfive,
  author    = {Goldberg, Lewis R.},
  title     = {{An alternative "description of personality": The Big-Five factor structure}},
  journal   = {Journal of Personality and Social Psychology},
  volume    = {59},
  number    = {6},
  pages     = {1216--1229},
  year      = {1990},
  month     = dec,
  doi       = {10.1037//0022-3514.59.6.1216},
  pmid      = {2283588},
  publisher = {American Psychological Association},
  address   = {United States},
}

@article{SchwartzArepository,
author = {Schwartz, Shalom},
year = {2021},
month = {09},
pages = {},
title = {{A Repository of Schwartz Value Scales with Instructions and an Introduction}},
volume = {2},
journal = {Online Readings in Psychology and Culture},
doi = {10.9707/2307-0919.1173}
}

@article{mcgrath2019via,
  title={{The VIA Assessment Suite for Adults: Development and Initial Evaluation Revised Edition}},
  author={McGrath, Robert E.},
  journal={Cincinnati, OH: VIA Institute on Character},
  year={2019}
}

@article{bruhlmann2024effectiveness,
  title={The effectiveness of warning statements in reducing careless responding in crowdsourced online surveys},
  author={Br{\"u}hlmann, Florian and Memeti, Zgjim and Aeschbach, Lena F. and Perrig, Sebastian A. C. and Opwis, Klaus},
  journal={Behavior research methods},
  volume={56},
  number={6},
  pages={5862--5875},
  year={2024},
  publisher={Springer},
  doi={https://doi.org/10.3758/s13428-023-02321-z}
}

@inproceedings{berg-kirkpatrick-etal-2012-empirical,
    title = "{An Empirical Investigation of Statistical Significance in {NLP}}",
    author = "Berg-Kirkpatrick, Taylor  and
      Burkett, David  and
      Klein, Dan",
    editor = "Tsujii, Jun{'}ichi  and
      Henderson, James  and
      Pa{\c{s}}ca, Marius",
    booktitle = "Proceedings of the 2012 Joint Conference on Empirical Methods in Natural Language Processing and Computational Natural Language Learning",
    month = jul,
    year = "2012",
    address = "Jeju Island, Korea",
    publisher = "Association for Computational Linguistics",
    url = "https://aclanthology.org/D12-1091/",
    pages = "995--1005"
}
\bibliographystyle{acl_natbib}

\clearpage

\appendix

\section*{Appendix}
\section{Survey Information and Dimensions of Traits}\label{sec:appendixA}
The full lists of traits corresponding to each theory are presented in
Table \ref{tab:trait_table}.

\paragraph{Big5.} 
The Big Five, also known as five-factor model (FFM), is based on the fundamental lexical hypothesis, which suggests that key personality traits become encoded in various languages \cite{goldberg1990bigfive}.
This theory presents five dimensions of personality: openness to experience, conscientiousness, extraversion, agreeableness, and neuroticism.
We obtain official items from the 50-item IPIP representation of the \citet{goldberg1992development} markers for the Big-Five factor structure \cite{goldberg1999broad}.\footnote{\url{https://ipip.ori.org/newNEODomainsKey.htm}}
This survey asks respondents to indicate how accurately the statement describes their typical behavior or attitudes. Responses are given on a 5-point Likert scale ranging from ``Very Accurate'' to ``Very Inaccurate''.

\paragraph{Schwartz.}

Schwartz's theory of basic values categorizes human values into 10 types \cite{schwartz1992universals}.
The Portrait Values Questionnaire (PVQ) was developed to measure these categories. We obtain official items from \citet{SchwartzArepository}. This survey asks respondents to indicate how much the people described in each item are like them. Responses are given on a 6-point Likert scale ranging from ``Very Much Like Me'' to ``Not at All Like Me''.

\paragraph{VIA.}
The Values in Action (VIA) framework categorizes personality strengths into 24 facets across six broad character strengths \cite{peterson2004character}. 
The VIA Inventory of Strengths – Mixed (VIA-IS-M) is a questionnaire designed to assess these facets \cite{mcgrath2019via}. 
We obtain official items from the VIA Institute on Character.\footnote{\url{https://www.viacharacter.org/}} This survey asks respondents to indicate whether the statement describes what they are like. Responses are given on a 5-point Likert scale ranging from ``Very much like me'' to ``Very much unlike me''.

\section{Construction of Initial Item Pools}\label{appendix:initial_item_pool}

\begin{figure}[h]\label{fig:item_generation_prompt}
    \centering
    \includegraphics[width=0.48\textwidth]{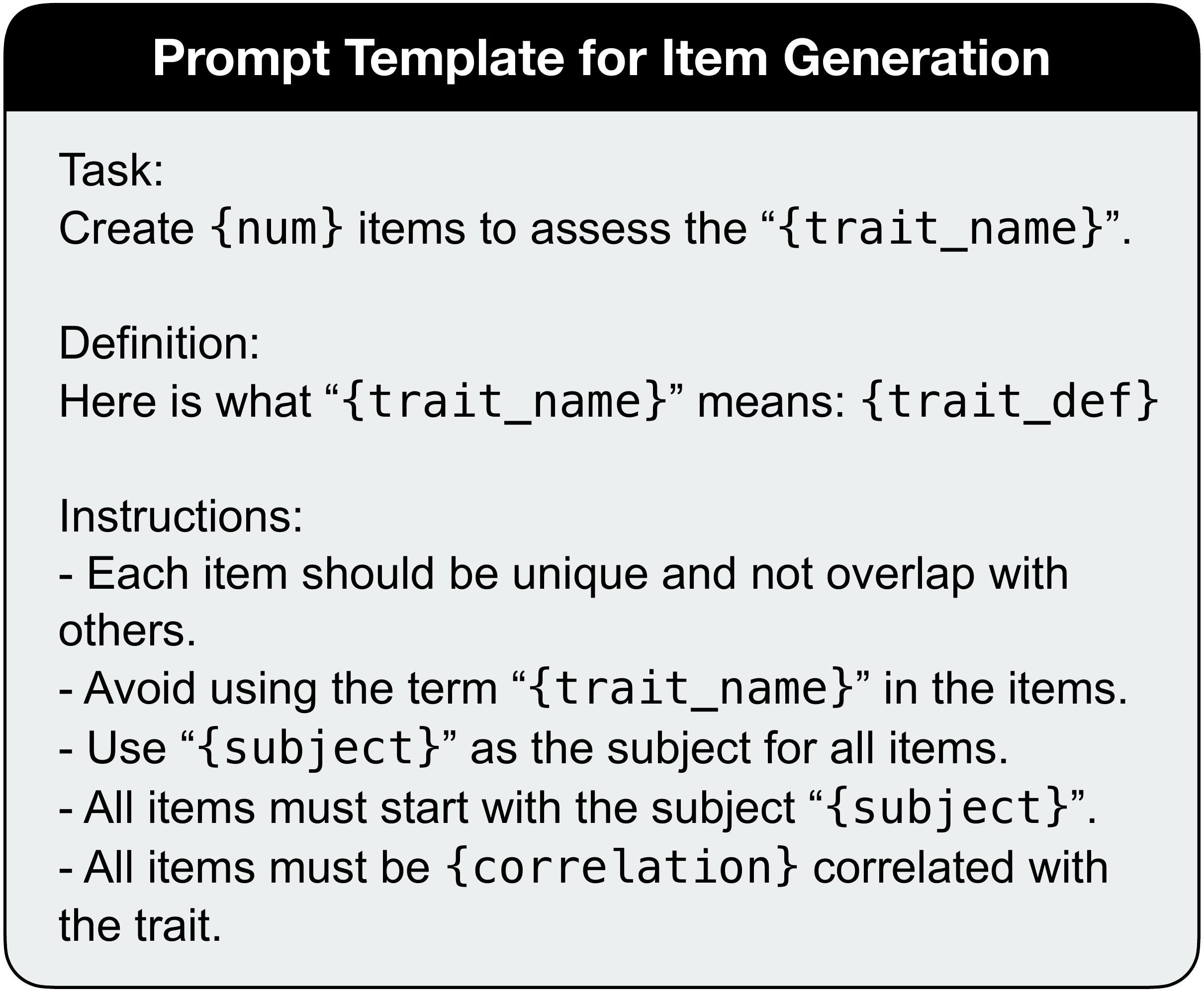}
    \caption{Prompt template used for item generation.}
    \label{fig:item_generation_prompt}
\end{figure}

In this section, we introduce the procedure of constructing an initial pool of psychometric items based on the framework proposed by \citet{yao2025clave}.

Initially, we used LLMs (with temperature set to 1) to generate a candidate pool of items four times the size of the official items. The template shown in Figure \ref{fig:item_generation_prompt} illustrates the five input fields and five instructions used in our study.
First, we supply definitions extracted from the official publications—while instructing the model not to use the exact trait definition within any item.
Second, we provide a grammatical subject identical to that used in the official items, requiring each item to begin with this subject.
Third, we specify the exact number of items (twice the number of official items) to be generated.
Fourth, we indicate the desired direction of correlation between generated items and the target trait, prompting separate generations for positively and negatively correlated items.
Finally, we enforced semantic uniqueness by instructing the model to produce non‐overlapping items. As a result, our procedure yielded four times as many candidate items as the official surveys.

Next, we refined this item pool by using a three-step embedding-and-clustering approach, building on recent value‐evaluation research demonstrating the effectiveness of combining LLMs with clustering‐based approaches \cite{yao2025clave}.
The first step is obtaining the semantic embeddings of items. We converted each item into a numeric vector embedding using OpenAI's text-embedding-3-small model. These embeddings capture the meaning of each item so that similar items are placed near one another in this high-dimensional space.
The second step is clustering. We applied K-means clustering to group these embeddings into roughly $n/4$ clusters, where $n$ is the total number of items. Each cluster represents a distinct semantic theme within the pool.
The final step is selecting representative items. From each cluster, we pick the item whose embedding lies closest to the cluster's centroid, assuming this item best represents the cluster's overall meaning.
As a result, this refinement process reduces the number of candidate items in the initial item pool to match the number of official items.

Through this process, we were able to generate new items covering a diverse range of situations.
While official surveys enable effective psychometric assessment, their items often include broad and abstract content (e.g., ``I am always prepared'' for measuring \textit{Conscientiousness} in the Big5).
In contrast, the items generated by LLMs tend to capture more concrete scenarios (e.g., ``I consistently meet deadlines for my assignments.'').

\section{Experiments with Different Trait Levels}\label{appdendix:low_experiments}

\begin{table}[t]\footnotesize
\begin{center}
\begin{tabular}{l@{\hskip 8pt}c@{\hskip 8pt}c@{\hskip 6pt}c@{\hskip 6pt}c@{\hskip 8pt}c@{\hskip 8pt}c}
\toprule
    \textbf{Method} & \multicolumn{4}{c}{\textbf{CV$\uparrow$}} & \textbf{DV$\downarrow$} & \textbf{ICR$\uparrow$} \\
    \cmidrule{2-7}
     & \scriptsize Score & \scriptsize Per. & \scriptsize NDCG & \scriptsize @$N$ & \scriptsize Score & \scriptsize Score \\
\midrule
    \multicolumn{7}{c}{Big5} \\
\midrule
    Low & .603 & 87.9 & .418 & .215 & \textbf{.294} & .890 \\
    Mixed & .623 & 94.1 & .529 & .385 & .297 & \textbf{.907} \\
    High & \textbf{.632} & \textbf{99.3} & \textbf{.568} & \textbf{.455} & \textbf{.294} & .904 \\
\midrule
    \multicolumn{7}{c}{Schwartz's Theory of Basic Values} \\
\midrule
    Low & .336 & 85.6 & \textbf{.697} & .475 & \textbf{.134} & \textbf{.767} \\
    Mixed & \textbf{.349} & \textbf{89.0} & .692 & \textbf{.517} & .140 & .736 \\
    High & .347 & 87.1 & .664 & .492 & .145 & .740 \\
\midrule
    \multicolumn{7}{c}{VIA} \\
\midrule
    Low & .583 & 86.0 & \textbf{.660} & \textbf{.488} & \textbf{.296} & .769 \\
    Mixed & .578 & 83.8 & .637 & .446 & .302 & .789 \\
    High & \textbf{.586} & \textbf{88.5} & .657 & .456 & .299 & \textbf{.803} \\
\bottomrule
\end{tabular}
\end{center}
\caption{Performance of selected items across different trait levels of the virtual respondents. For mediators, we employ the best-performing mediator generation strategies from Section \ref{sec:results}: Trait (free) for Big5 and VIA, and Trait (CAPS) for Schwartz. (Per.: percentile, NDCG: NDCG@All, @$N$: NDCG@$N$).}
\label{appdendix:trait_level}
\end{table}

In our main experiments, the prompt for each virtual respondent begins with a target trait set to a high level (e.g., ``I highly value \{trait\}'' in Figure~\ref{fig:simulation_prompt}).
Here, we explore the effectiveness of including other trait levels in virtual respondents.
To create virtual respondents with low trait levels, we modify the opening sentence to ``I oppose \{trait\}''.
In this case, we need mediators that may facilitate behaviors related to the trait.
Although we attempted to generate such mediators using the same mediator generation strategies in \ref{sec:mediator_generation}, many of the generated mediators undesirably leaned toward suppressing the trait. 
To address this, we opted to reverse the original mediators generated in our main experiments, using GPT-4.1. 
We keep the rest of the virtual respondent prompt unchanged.
We conduct two experiments: (1) 500 virtual respondents with \textbf{low} trait levels only, and (2) 500 virtual respondents with \textbf{mixed} trait levels (250 low and 250 high).

As shown in Table \ref{appdendix:trait_level}, experiments conducted only with high-trait virtual respondents achieve the highest CV scores for Big5 and VIA.
For Schwartz, the high setting also records high scores with marginal underperformance compared to the mixed setting.
Based on these results, we conclude that setting virtual respondents' trait levels to high produces the most stable and superior performance, leading us to adopt it as our main experimental configuration.

\section{Details of Human Survey}\label{appendix:human_survey}
Since all our surveys are constructed in English, we recruit adults who are native English speakers or fluent in English through the online survey platform Prolific.
To recruit reliable human respondents, we target individuals who have completed more than 100 previous submissions on this platform and maintain an approval rate of 98\% or higher.
We recruit participants with equal gender distribution and equal numbers across age groups specified in Table \ref{tab:human_respondents_table}.
A total of 339 respondents were recruited, and 307 respondents were finally included in our analysis based on the attention check procedure outlined in Appendix \ref{appendix:attention_check}.

\subsection{Demographics of Human Respondents}\label{appendix:demo_human}

% Table: Demographics of human respondents
\begin{table}[t]\footnotesize
\centering
\setlength{\tabcolsep}{3pt}
\begin{tabularx}{\linewidth}{
    >{\centering\arraybackslash}m{1.7cm} |
    >{\centering\arraybackslash}m{1cm} |
    >{\centering\arraybackslash}m{2.2cm} |
    >{\centering\arraybackslash}m{1.8cm}
}
\toprule
    \textbf{Survey} & \textbf{N} & \textbf{Gender} & \textbf{Age} \\
\midrule
    Big5-G & \centering75 & - Male: 37\newline- Female: 38 & 
- 20\textasciitilde29: 16 \newline
- 30\textasciitilde39: 14 \newline
- 40\textasciitilde49: 15 \newline
- 50\textasciitilde59: 15 \newline
- 60\textasciitilde100: 15 \\
\midrule
    PVQ & \centering76 & - Male: 37\newline- Female: 39 & 
- 20\textasciitilde29: 15 \newline
- 30\textasciitilde39: 15 \newline
- 40\textasciitilde49: 16 \newline
- 50\textasciitilde64: 16 \newline
- 60\textasciitilde100: 14 \\
\midrule
    VIA-IS-M\newline(Part 1) & \centering80 & - Male: 40\newline- Female: 38 \newline- Non-binary / Third gender: 2 &
- 20\textasciitilde29: 16 \newline
- 30\textasciitilde39: 18 \newline
- 40\textasciitilde49: 15 \newline
- 50\textasciitilde59: 16 \newline
- 60\textasciitilde100: 15 \\
\midrule
    VIA-IS-M\newline(Part 2) & \centering76 & - Male: 37\newline- Female: 38\newline- Non-binary / Third gender: 1 &
- 20\textasciitilde29: 16 \newline
- 30\textasciitilde39: 16 \newline
- 40\textasciitilde49: 14 \newline
- 50\textasciitilde59: 15 \newline
- 60\textasciitilde100: 15 \\
\bottomrule
\end{tabularx}
\caption{Demographics of human respondents.}
\label{tab:human_respondents_table}
\end{table}

The demographic information of the final respondents is presented in Table~\ref{tab:human_respondents_table}.

\subsection{Attention Check Procedure}\label{appendix:attention_check}
To ensure human respondents' attentiveness, we employ attention-check items consisting of two types: 1) duplicate items that repeat previously presented items, and 2) bogus items that require consistently negative responses.
The attention-check items are distributed across three pages, with each page containing one bogus item and one duplicate item.

\paragraph{Duplicate Items.}
Duplicate items are randomly selected from the official items and presented repeatedly in the survey.
Answers from their initial appearance are used to assess the respondent's trait scores, while answers from the second appearance serve only as attention checks.
Respondents who show a discrepancy of two or more points on the Likert scale on more than two occasions are considered inattentive and excluded from the analysis.

\paragraph{Bogus Items.}
Bogus items ask respondents about highly improbable actions or behaviors with which they are not expected to agree \citep{bruhlmann2024effectiveness}.
We generate three bogus items for each survey using GPT-4o with the prompt below.
An example bogus item generated by this prompt is ``I can speak every language in the world fluently''.
When a respondent answers positively to these items, their task is immediately terminated.

\section{Details of the Human Evaluation of Mediators}\label{appendix:mediator_evaluation}
The mediator categories and their definitions are presented in Table~\ref{tab:mediator_category}.

\begin{tcolorbox}[title=Prompt used for Bogus Item Generation, halign=left, boxrule=0.5pt]\small

Task: \\
Create 12 bogus items to detect careless responses in an online survey. \\
\vspace{2ex}
Instructions: \\
- Use ``I'' as the subject for 9 items and use ``They'' as the subject for 3 items. \\
- Each item should be unique and not overlap with others. \\
\end{tcolorbox}

\section{Definitions of Evaluation Metrics}\label{appendix:denfinition_percentile_ndcg}

\paragraph{Percentile.}
We compute the percentile to assess the relative quality of the selected item set. The percentile of the selected item set $s^*$ is defined as:
\begin{equation}
\begin{split}
&\text{Percentile}(s^*) \;=\; \\
&\frac{1}{|S|}\sum_{s \in S} \mathbb{I}\!\left[\; \mathrm{CV}(s) \le \mathrm{CV}(s^*) \;\right] \times 100
\end{split}
\end{equation}
where $S$ is a set of all possible $N$-item combinations from the initial item pool, $\mathbb{I}[\cdot]$ is the indicator function, and $\text{CV}(s)$ is the CV score of the item set $s$.

\paragraph{NDCG and NDCG@$N$.}
To measure how accurately each mediator generation strategy ranks individual items, we use the normalized discounted cumulative gain (NDCG). It is based on the discounted cumulative gain (DCG), defined as:
\begin{equation}
\mathrm{DCG} \;=\; \sum_{i=1}^{M} \frac{2^{\mathrm{rel}_i}}{\log_2(i+1)}, 
\end{equation}
where $\mathrm{rel}_i$ denotes the relevance (here, the ranking of the item based on CV scores) of the item at rank $i$, and $M$ is the total number of items.
The ideal DCG (IDCG) is computed by sorting items in descending order of their true CV scores, denoted $\mathrm{rel}^\star_i$. The normalized score is then given by:
\begin{equation}
\mathrm{NDCG} \;=\; \frac{\mathrm{DCG}_M}{\mathrm{IDCG}_M} \in [0,1]
\end{equation}
which always lies in the range [0,1], with higher values indicating more accurate rankings.

In practice, a truncated version known as NDCG@$k$ is widely used, where the parameter $k$ specifies the number of top-ranked items considered.
In our evaluation, we report both NDCG, obtained by setting $k=M$ to the total number of items to assess full-ranking accuracy, and NDCG@$N$, obtained by setting $k=N$ to assess ranking accuracy within the top-$N$ items of the selected set.

\section{Prompt Templates for Mediator Generation}\label{appendix:mediator_generation}

\subsection{Trait (Free)}\label{appendix:mediator_trait}

\begin{tcolorbox}[title=Freely generate mediators with traits, halign=left, boxrule=0.5pt]\small

Trait: \texttt{\{trait\_name\}} \\
Definition: \texttt{\{trait\_def\}} \\
\vspace{2ex}
List possible human characteristics or backgrounds that would be unlikely 
or contradictory for someone who strongly values \texttt{\{trait\_name\}}. 
Just number them without detailed explanation. Make many values as possible.

\end{tcolorbox}

\subsection{Trait (CAPS)}\label{appendix:mediator_caps}
The italicized sections in the prompt below represent the five categories of mediators proposed by the CAPS theory \cite{mischel1995cognitive}. We introduce each category one at a time to separately generate mediators across five iterations.

\begin{tcolorbox}[title=Generate mediators based on CAPS, halign=left, boxrule=0.5pt]\small
Trait: \texttt{\{trait\_name\}} \\
Definition: \texttt{\{trait\_def\}} \\
\vspace{2ex}
List possible \texttt{\{}\textit{(1) Situation Encodings, (2) Expectancies and Beliefs, (3) Affective Responses, (4) Goals and Values, (5) Competencies and Self-regulatory Plans}\texttt{\}} that could create internal conflict or lead to changes in behavior for someone who strongly values \texttt{\{trait\_name\}}. List as many items as possible. Number each item without detailed explanation.
\end{tcolorbox}

\subsection{Trait+Item}\label{appendix:mediator_item}
We use generated items to create diverse mediators. The official items from the original surveys are not used in this process.

\begin{tcolorbox}[title=Generate mediators with an item, halign=left, boxrule=0.5pt]\small
Trait: \texttt{\{trait\_name\}} \\
Definition: \texttt{\{trait\_def\}} \\
Survey Item: \texttt{\{item\}} \\
\vspace{2ex}
Generate a single personal characteristic, background factor, or life circumstance that could plausibly lead someone—even among people who highly value \texttt{\{trait\_name\}}—to respond contrary to the survey item above.
\end{tcolorbox}

For the mediators generated by Trait (Free) (\ref{appendix:mediator_trait}), Trait (CAPS) (\ref{appendix:mediator_caps}), and Trait+Item (\ref{appendix:mediator_item}), we convert them into persona sentences using GPT-4.1 before integrating them with persona profiles.
The prompt is as follows:

\begin{tcolorbox}[title=Convert mediators into persona sentences, halign=left, boxrule=0.5pt]\small
For each of the contents below, write a single sentence introducing a person's persona. Use ``I'' as a subject. \\
\vspace{2ex}
\texttt{\{List of mediators\}}
\end{tcolorbox}

\subsection{Trait+WVS}\label{appendix:mediator_wvs}
The World Values Survey (WVS) is a globally comprehensive survey that examines changing human beliefs and values and their impact on social and political life across multiple countries worldwide.
The survey encompasses questions about social, economic, religious, and ethical values, including topics such as the importance of family in life.
Based on these question topics and sentences, we create persona sentences assuming individuals who agree with these questions (Step 1) and use the LLM to determine whether these personas conflict with the traits (Step 2).
Sentences classified as conflicting values serve as mediators for the corresponding traits.

\begin{tcolorbox}[title=Step 1: Convert WVS questions into persona sentences, halign=left, boxrule=0.5pt]\small
Create a concise persona sentence who say `Yes/Agree/Likely' to the question below. Use ``I'' as the subject. \\
\vspace{2ex}
Question Topic: \texttt{\{WVS\_question\_topic\}} \\
Question: \texttt{\{WVS\_question\_sentence\}} \\
\end{tcolorbox}

\begin{tcolorbox}[title=Step 2: Determine whether the given persona sentences conflict with each trait, halign=left, boxrule=0.5pt]\small
Consider the given values and the personality trait below. \\
Determine whether the given values conflict with the personality trait, making it difficult for individuals to respond accurately to questions designed to measure the trait. \\
\vspace{2ex}
<Personality Trait> \\
\texttt{\{trait\_name\}}: \texttt{\{trait\_def\}} \\
\vspace{2ex}
<Values> \\
\texttt{\{persona\_sentence\}}
\end{tcolorbox}

\section{Prompt Templates for Mediator-Guided Simulation}

\subsection{Big5}
\begin{tcolorbox}[title=Prompt Template for Big5-G Simulation, halign=left, boxrule=0.5pt, breakable]\small
\texttt{\{Target Trait + Mediator-integrated persona profile\}}

\vspace{2ex}
<Instruction> \\
Based on all the information provided above, select only one answer from the <Answer Choices> to indicate how accurately the <Statement> describes this person’s typical behavior or attitudes. \\
\vspace{2ex}
<Statement> \\
\texttt{\{Item for target trait\}} \\

\vspace{2ex}
<Answer Choices>\\
\text{[}Very Accurate, Moderately Accurate, Neither inaccurate nor accurate, Moderately Inaccurate, Very Inaccurate\text{]}
\end{tcolorbox}

\subsection{Schwartz}
\begin{tcolorbox}[title=Prompt Template for PVQ Simulation, halign=left, boxrule=0.5pt]\small
\texttt{\{Target Trait + Mediator-integrated persona profile\}}

\vspace{2ex}
<Instruction> \\
Based on all the information provided above, select only one answer from the <Answer Choices> to indicate the degree to which this person is like them, as described in the <Statement>. \\
\vspace{2ex}
<Statement> \\
\texttt{\{Item for target trait\}} \\

\vspace{2ex}
<Answer Choices>\\
\text{[}Very Much Like Them, Like Them, Somewhat Like Them, A Little Like Them, Not Like Them, Not Like Them at All\text{]}
\end{tcolorbox}

To allow the simulation LLM to understand the task better, we replaced the pronouns in the answer choices in the official survey from the first-person pronoun ``Me'' to the third-person pronoun ``Them''.

\subsection{VIA}
\begin{tcolorbox}[title=Prompt Template for VIA-IS-M Simulation, halign=left, boxrule=0.5pt]\small
\texttt{\{Target Trait + Mediator-integrated persona profile\}}

\vspace{2ex}
<Instruction> \\
Based on all the information provided above, select only one answer from the <Answer Choices> to indicate the degree to which the <Statement> describes what the person is like. \\
\vspace{2ex}
<Statement> \\
\texttt{\{Item for target trait\}} \\

\vspace{2ex}
<Answer Choices>\\
\text{[}Very Much Like Them, Like Them, Somewhat Like Them, A Little Like Them, Not Like Them, Not Like Them at All\text{]}
\end{tcolorbox}

To allow the simulation LLM to understand the task better, we replaced the pronouns in the answer choices in the official survey from the first-person pronoun ``Me'' to the third-person pronoun ``Them''.

\section{Baseline}\label{appendix:baseline}

\subsection{LLM-as-a-Judge}
\begin{tcolorbox}[title=Prompt Template used for LLM-based Item Evaluation, halign=left, boxrule=0.5pt, breakable]

Please evaluate the following psychometric item according to the criteria below. \\
\vspace{2ex}
Item: ``\texttt{\{item\}}'' \\
Target Trait: \texttt{\{trait\_name\}} \\
Trait Definition: \texttt{\{trait\_def\}} \\
Expected Correlation (direction): \texttt{\{correlation\}} \\
\vspace{2ex}
<Evaluation criteria> \\
1. Validity \\
a) Convergent validity - the degree to which the item accurately measures the target trait. \\
b) Discriminant validity - the degree to which the item does not substantially measure traits other than the target trait. \\
\vspace{2ex}
2. Reliability \\
a) Test-retest reliability - the degree of consistency in responses when the item is administered to the same individual at different times. \\
\vspace{2ex}
<Response format> \\
(Use a 1–100 scale, where 1 = very poor and 100 = excellent.): \\
Convergent validity: [score 1-100] \\
Discriminant validity: [score 1-100] \\
Test-retest reliability: [score 1-100] \\
Explanation: [brief rationale for each rating] \\
\end{tcolorbox}

\section{Significance Tests}\label{appendix:significance_test}

\begin{figure}[!t] 
    \centering
    \includegraphics[width=0.43\textwidth]{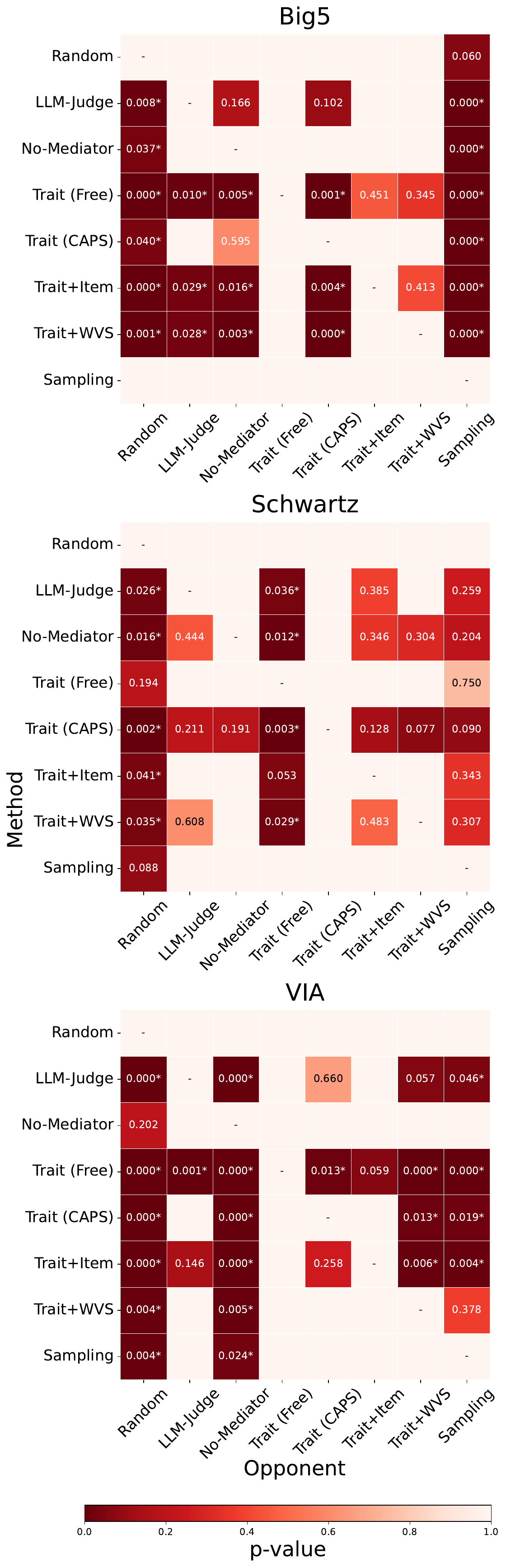}
    \caption{Results of significance test for Section~\ref{sec:results}. All reported p-values are based on a directional hypothesis of $y>x$; comparisons in the opposite direction ($y<x$) are omitted. An asterisk (*) indicates p-value $< 0.05$.}
    \label{fig:significance_test}
\end{figure}

To determine whether the observed results presented in Section~\ref{sec:results} and \ref{sec:simulation_components} are statistically meaningful, we conduct significance testing.
\citet{berg-kirkpatrick-etal-2012-empirical} proposed a recentered paired bootstrap estimator of the one-sided p-value: repeatedly draw bootstrap resamples $x^{(i)}$ from the full test set $x$ (with replacement), compute the metric difference $\delta(x^{(i)})= m(A, x^{(i)})-m(B,x^{(i)})$ between systems $A$ and $B$, where $m()$ denotes the evaluation metric, and approximate $p\approx s/b$ where $s$ counts resamples with $\delta(x^{(i)})>2\delta(x)$ and $b$ is the number of resamples.

Following this methodology, our significance test proceeds as follows.
We first select hypotheses in which system A outperformed system B in terms of CV score in Section~\ref{sec:results} and \ref{sec:simulation_components}.
Then, we resample the initial item pool 1,000 times with replacement for each trait, maintaining the original pool size.
The top $N$ items are selected for each item selection method.
Next, we compute the difference in CV scores for each pair of methods and compare it with the difference on the original test set.
We then estimate the p-value by applying the criterion $\delta(x^{(i)}) > 2\delta(x)$.

\subsection{Main Results}\label{appendix:significance_test_main}
The results in Figure~\ref{fig:significance_test} indicate that the findings in Section~\ref{sec:results} remain statistically stable, whereas the outcomes for Schwartz are less consistent.
The best methods for Big5 and VIA (Trait (Free)) consistently outperform all baseline approaches and the Sampling method, indicating that our simulation framework yields more robust and reliable performance differences across stochastic runs.
In particular, these best methods outperformed the No-Mediator setting with $p < 0.05$, underscoring the importance of the mediator in identifying valid items.
However, for Schwartz, it was difficult to identify consistent results.
One plausible explanation is that this result stems from the weaker validity of the Schwartz items.
As shown in Table~\ref{tab:result_table}, the overall CV scores of Schwartz items are substantially lower than those for Big5 and VIA.
Unlike other theories, Schwartz theory defines a circular structure in which positively correlated values (e.g., universalism and benevolence) are placed closely.
Due to these inherent correlations between traits, it is more difficult to generate high-validity items, and methods may also struggle to distinguish higher- from lower-quality items.

\subsection{Simulation Components}\label{appendix:significance_test_ablation}

\begin{figure}[!t] 
    \centering
    \includegraphics[width=0.43\textwidth]{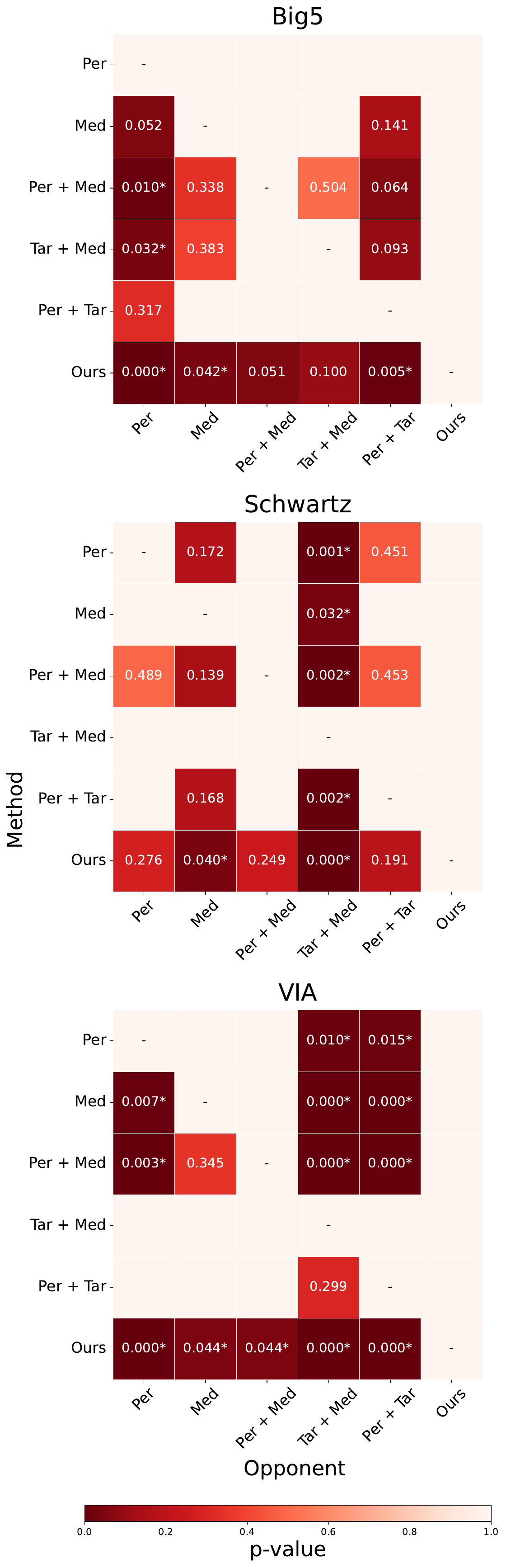}
    \caption{Results of significance test for Section \ref{sec:simulation_components}. All reported p-values are based on a directional hypothesis of $y>x$; comparisons in the opposite direction ($y<x$) are omitted. An asterisk (*) indicates p-value $< 0.05$.}
    \label{fig:significance_test_ablation}
\end{figure}

The results in Figure~\ref{fig:significance_test_ablation} show that our full setting (i.e., \textit{Ours}) consistently outperforms all other combinations.
Notably, more consistent improvements are observed for Big5 and VIA, while Schwartz exhibits comparatively less consistent results.
This trend aligns with the findings in Appendix~\ref{appendix:significance_test_main}.
These findings indicate that all components contribute meaningfully to performance, with the full integration yielding the most reliable improvements.

\section{Maximum DV Scores}\label{appendix:dv_max}

\begin{table}[!h]\small
\begin{center}
\begin{tabular}{l|ccc}
\toprule
    Method       & Big5  & Schwartz & VIA   \\ 
\midrule
    Random       & .594 & .398    & .578 \\
    LLM-Judge    & \underline{.546} & .399    & .587 \\
    No-Mediator  & .552 & .399    & \textbf{.549} \\
    Trait (Free) & .568 & \textbf{.363}    & .594 \\
    Trait (CAPS) & .567 & .411    & .580 \\
    Trait+Item   & .557 & \underline{.383}    & \underline{.573} \\
    Trait+WVS    & .585 & .385    & .584 \\
    Sampling     & \textbf{.530} & .387    & .581 \\
\midrule
    Oracle    & .593 & .396 & .585 \\
    Official     & .624 & .453 & .573 \\
\bottomrule
\end{tabular}
\caption{Maximum DV scores for each survey.}
\label{tab:dv_max}
\end{center}
\end{table}

As a reference, we provide the maximum DV scores in Table~\ref{tab:dv_max}.
For each trait, we first identify the worst DV score, defined as the maximum Spearman correlation with non-target traits among the top $N$ items selected by each method.
Each cell then shows the average of these values over all traits.

The results show that the differences across methods are small, at most around 0.03 to 0.05.
We also recommend interpreting the maximum DV score together with the CV score rather than on its own, as illustrated by the ratio of DV to CV.

\section{Ablation Studies on Item Generation Scale}\label{appendix:scale_ablation}

% Table: Ablation Studies on Item Generation Scale
\begin{table}[!t]\footnotesize
\begin{center}
\begin{tabular}{l@{\hskip 8pt}c@{\hskip 8pt}c@{\hskip 6pt}c@{\hskip 6pt}|c@{\hskip 8pt}c@{\hskip 8pt}c}
\toprule
    \textbf{Method} & \multicolumn{3}{c|}{\textbf{Scale $=2$}} & \multicolumn{3}{c}{\textbf{Scale $=3$}} \\
    \cmidrule{2-7}
     & \scriptsize CV$\uparrow$ & \scriptsize DV$\downarrow$ & \scriptsize ICR$\uparrow$ & \scriptsize CV$\uparrow$ & \scriptsize DV$\downarrow$ & \scriptsize ICR$\uparrow$ \\
\midrule
    \multicolumn{7}{c}{Big5} \\
\midrule
    Random & .545 & \underline{.276} & .850 & .545 & \underline{.276} & .850 \\
    LLM-Judge & .583 & .284 & .882 & .595 & .288 & .890 \\
    No-Mediator & .555 & .282 & .867 & .573 & .291 & .885 \\
    Trait (Free) & \textbf{.592} & .287 & \textbf{.886} & \textbf{.620} & .292 & \textbf{.900} \\
    Trait (CAPS) & .561 & .283 & .864 & .579 & .290 & .886 \\
    Trait+Item & \underline{.590} & .289 & .874 & \underline{.615} & .291 & .889 \\
    Trait+WVS & .565 & .283 & \underline{.877} & .592 & .297 & \underline{.896} \\
    Sampling & .538 & \textbf{.272} & .849 & .525 & \textbf{.270} & .846 \\
\midrule
    \multicolumn{7}{c}{Schwartz's Theory of Basic Values} \\
\midrule
    Random & .234 & \textbf{.134} & .533 & .234 & \textbf{.134} & .533 \\
    LLM-Judge & .292 & .137 & .646 & .316 & .140 & .678 \\
    No-Mediator & .289 & .138 & .678 & .319 & .138 & \underline{.745} \\
    Trait (Free) & .294 & \textbf{.134} & \textbf{.700} & .313 & .139 & \underline{.745} \\
    Trait (CAPS) & \underline{.295} & .138 & .683 & \textbf{.328} & .143 & .740 \\
    Trait+Item & \textbf{.302} & .138 & .647 & .319 & .140 & .684 \\
    Trait+WVS & .293 & \underline{.136} & \underline{.689} & \underline{.322} & \underline{.136} & \textbf{.749} \\
    Sampling & .264 & .138 & .611 & .280 & .138 & .653 \\
\midrule
    \multicolumn{7}{c}{VIA} \\
\midrule
    Random & .500 & \textbf{.280} & .674 & .500 & \textbf{.280} & .674 \\
    LLM-Judge & \underline{.545} & .286 & \textbf{.740} & .564 & .290 & \underline{.772} \\
    No-Mediator & .507 & .286 & .695 & .506 & \underline{.284} & .705 \\
    Trait (Free) & .543 & .292 & \underline{.737} & \textbf{.566} & .297 & \textbf{.778} \\
    Trait (CAPS) & .534 & .294 & .729 & .547 & .298 & .756 \\
    Trait+Item & \textbf{.547} & \underline{.281} & .727 & \underline{.568} & \textbf{.280} & .748 \\
    Trait+WVS & .524 & .298 & .727 & .527 & .300 & .743 \\
    Sampling & .515 & .287 & .703 & .524 & .288 & .713 \\
\bottomrule
\end{tabular}
\end{center}
\caption{Results of the ablation studies on item generation scale (with scales 2 and 3). \textbf{Bold}: best performance, \underline{Underlined}: second-best performance.}
\label{tab:ablation_scale_table}
\end{table}

To examine the robustness of our results with respect to the item generation scale, we conducted an ablation study with scales of 2 and 3, in addition to the original scale of 4.
We constructed item pools of each scale by subsampling without replacement from the original scale, repeating this procedure 1,000 times.
For each subsample, we selected the top $N$ items using each method and computed the CV, DV, and ICR scores.
We report the mean scores over all subsamples in Table~\ref{tab:ablation_scale_table}.

The results show that Trait (Free), which achieves the best CV score in the main results (Table~\ref{tab:result_table}), maintains the best CV and ICR scores across all scales for the Big5, and the second-best CV score for VIA at scale = 3.
For Schwartz, Trait (CAPS), the method with the best CV score in the main results, ranks best at scale = 3 and second-best at scale = 2.
Overall, these findings suggest a largely consistent pattern across different item generation scales.

% Table: Mediator Categories
\begin{table}[!t]\footnotesize
\centering
\begin{tabularx}{\linewidth}{>{\raggedright\arraybackslash}p{0.2\linewidth}X}
\toprule
    \textbf{Category} & \textbf{Definition} \\
\midrule
    Beliefs and Values & Core principles, convictions, and guiding ideals that influence thought, judgment, and behavior. This includes both abstract philosophical beliefs and personally significant moral or ethical values. \\
\midrule
    Emotions and Feelings & Subjective affective experiences, including moods, emotional states, and physiological responses associated with these experiences. \\
\midrule
    Habits and Behaviors & Recurrent actions, routines, and behavioral patterns that characterize an individual’s daily life, reflecting both voluntary and automatic tendencies. \\
\midrule
    Preferences and Interests & Likes, dislikes, and areas of personal engagement that reflect subjective inclinations, tastes, and attentional priorities. These may involve enjoyment of specific activities or objects. \\
\midrule
    Self-Concept and Abilities & One’s perceptions, evaluations, and beliefs about personal traits, skills, capacities, and potential for action; includes self-knowledge about competencies and identity. \\
\midrule
    Roles and Memberships & Positions, responsibilities, and affiliations within social, organizational, or community structures that influence identity and behavioral expectations. \\
\midrule
    Others’ Perceptions of Me & How one is perceived, described, or evaluated by other people. This encompasses reputations, social feedback, and interpersonal impressions that can shape self-concept and guide behavior. \\
\midrule
    Environment and Context & External circumstances, situational factors, and physical or social settings that shape experiences, behavior, and decision-making. \\
\midrule
    Other & Miscellaneous aspects not covered by the above categories. \\
\bottomrule
\end{tabularx}
\caption{Categories for mediators and their definitions.}
\label{tab:mediator_category}
\end{table}

\end{document}